\documentclass{article}

\usepackage[preprint]{corl_2025}

\usepackage{etoolbox} 
\usepackage[utf8]{inputenc} 
\usepackage[T1]{fontenc}    
\usepackage{fonttable}
\usepackage[english]{babel} 
\usepackage[protrusion=true,expansion=true]{microtype}
\usepackage{comment}
\usepackage{cancel}
\usepackage{csquotes}
\usepackage{listings}
\usepackage{nameref}
\usepackage{paralist}
\usepackage{xspace}
\usepackage{setspace}
\usepackage{makeidx}
\usepackage{pdfpages}
\usepackage{etoc} 
\usepackage[dvipsnames,table]{xcolor}

\usepackage{times}
\usepackage{microtype}

\usepackage{amsmath,amssymb} 
\usepackage{amsfonts}       
\usepackage{mathrsfs} 
\usepackage{mathtools}
\usepackage{array} 
\usepackage{nicefrac}       
\usepackage{breqn}          
\usepackage{textcomp}
\usepackage{bm} 
\usepackage{pifont}
\usepackage[
  separate-uncertainty = true,
  multi-part-units = repeat
]{siunitx}

\usepackage{graphicx}
\usepackage{wrapfig}
\usepackage{adjustbox} 
\usepackage{epsfig}

\usepackage{float}
\usepackage{stfloats}

\usepackage{placeins}
\usepackage{subcaption}
\usepackage[font=small,labelfont=bf]{caption}
\usepackage{lscape}      

\usepackage{tikz}  
\usepackage{makecell}
\usepackage{overpic}
\usepackage{algorithm}
\usepackage[noend]{algpseudocode}

\usepackage{array} 
\usepackage{tabularx}
\usepackage{multirow}
\usepackage{multicol}
\usepackage{booktabs}   
\usepackage{tablefootnote}

\usepackage{paralist}
\usepackage{enumitem}
\setlist[itemize]{noitemsep,leftmargin=*,topsep=0in}
\setlist[enumerate]{noitemsep,leftmargin=*,topsep=0in}

\usepackage{url}            
\PassOptionsToPackage{table}{xcolor}
\usepackage{color,colortbl} 
\usepackage{soul}

\PassOptionsToPackage{capitalize, noabbrev}{cleveref}

\hypersetup{pagebackref=false,breaklinks=true,colorlinks,urlcolor=blue,citecolor=blue,linkcolor=blue,bookmarks=false}

\usepackage{blindtext}
\usepackage{bbm}

\usepackage{lipsum}

\usepackage{titlesec}
\titlespacing{\section}{0pt}{0.25\baselineskip}{0.2\baselineskip}
\titlespacing{\subsection}{0pt}{0.15\baselineskip}{0.1\baselineskip}
\titlespacing{\subsubsection}{0pt}{0.05\baselineskip}{0.03\baselineskip}

\renewcommand{\paragraph}[1]{\vspace{0.2em}\noindent\textit{#1} --}

\newcommand*\algname{Adaptive 3D Scene Representation}
\newcommand*\algabbr{Adapt3R}

\title{\algabbr: Adaptive 3D Scene Representation \\ for Domain Transfer in Imitation Learning}

\author{Albert Wilcox$^{12}$, Mohamed Ghanem$^{1}$, Masoud Moghani$^{3}$, \\ \textbf{Pierre Barroso$^{2}$, Benjamin Joffe$^{12}$, Animesh Garg$^{1}$} \\
\vspace{0.03in}\\
$^1$Georgia Institute of Technology\quad$^2$Georgia Tech Research Institute\quad$^3$University of Toronto 
\\
}

\begin{document}
\maketitle

\begin{center}
    \centering
    \captionsetup{type=figure}
     \includegraphics[width=1.0\textwidth]{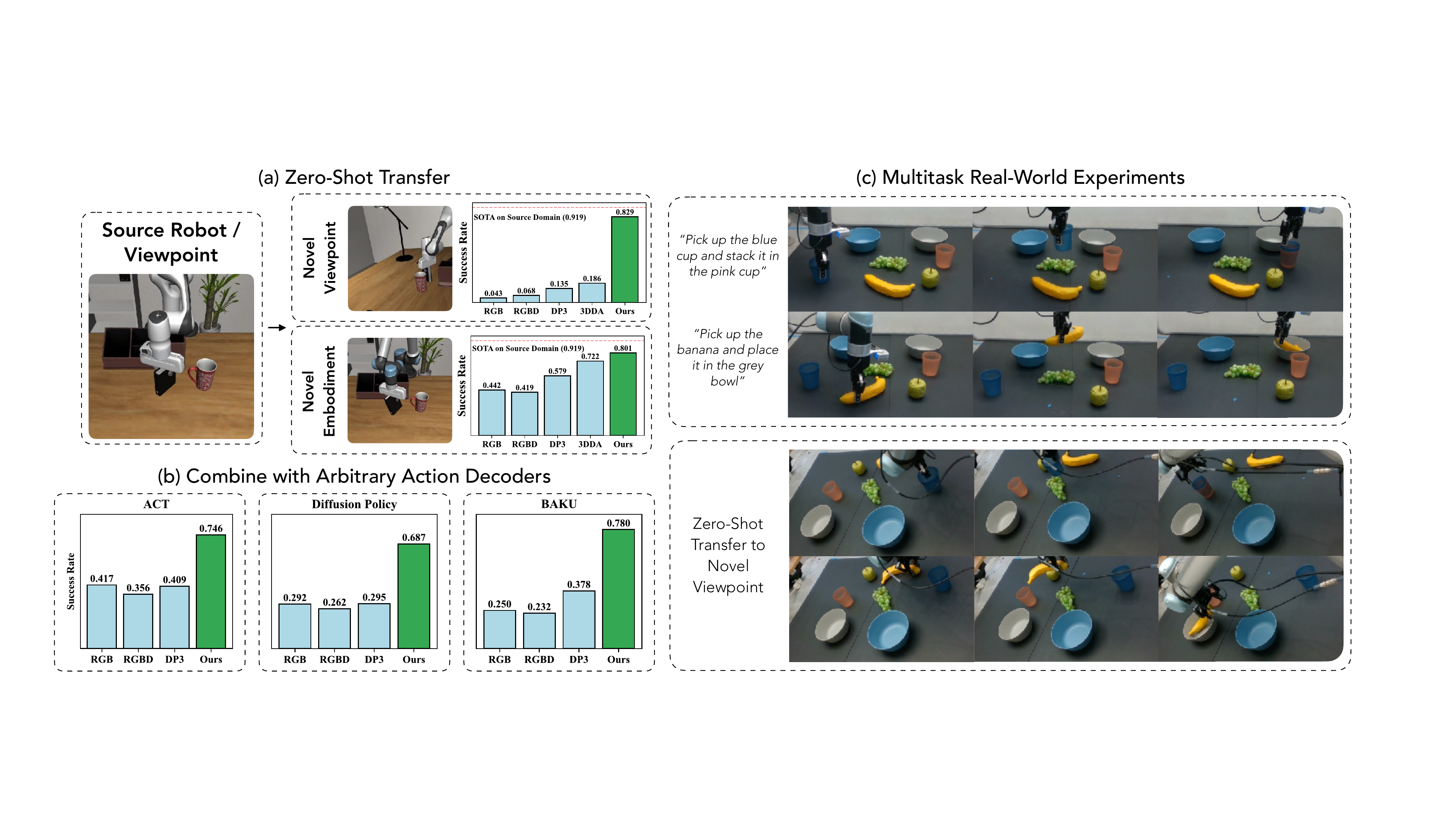}
    \caption{
    (a) \algabbr{} facilitates zero-shot transfer to novel embodiments and viewpoints. (b) \algabbr{} can be trained end-to-end as the encoder for a wide variety of imitation learning algorithms. (c) In a real-world multitask imitation learning benchmark, \algabbr{} enables zero-shot transfer to an unseen camera pose. 
    }
    \label{fig:teaser}
\end{center}

\begin{abstract}
    Imitation Learning can train robots to perform complex and diverse manipulation tasks, but learned policies are brittle with observations outside of the training distribution. 
    3D scene representations that incorporate observations from calibrated RGBD cameras have been proposed as a way to mitigate this, but in our evaluations with unseen embodiments and camera viewpoints they show only modest improvement. 
    To address those challenges, we propose \algabbr{}, a general-purpose 3D observation encoder which synthesizes data from calibrated RGBD cameras into a vector that can be used as conditioning for arbitrary IL algorithms. The key idea is to use a pretrained 2D backbone to extract semantic information, using 3D only as a medium to localize this information with respect to the end-effector.
    We show across 93 simulated and 6 real tasks that when trained end-to-end with a variety of IL algorithms, \algabbr{} maintains these algorithms' learning capacity while enabling zero-shot transfer to novel embodiments and camera poses. 
    For more results, visit \url{https://pairlab.github.io/Adapt3R}.
\end{abstract}


\section{Introduction}

An important goal of the robot learning community is to train generalist agents that can solve a wide variety of tasks, as others have done for computer vision \citep{deng2009imagenet, radford2021learning, he2022masked, oquab2023dinov2, ramesh2021zero, rombach2022high, kirillov2023segment} and natural language processing (NLP) \cite{radford2018improving,radford2019language,brown2020language,achiam2023gpt}.
While recent work in multitask imitation learning has achieved impressive results when evaluated on in-distribution domains \cite{rt-1, rt-2, shridhar2022peract, goyal2023rvt, ha2023scalingup, mete2024quest, baku}, these policies are notoriously brittle when faced with changes in factors such as camera pose, and are not able to reliably transfer to new robot embodiments. Even large scale data collection efforts in robotics \cite{vuong2023open, walke2023bridgedata, khazatsky2024droid} yield datasets orders of magnitude smaller than those used for vision and NLP that are so diverse and sparse that it's difficult to learn unified representations across them. Ultimately, collecting enough data to cover the full deployment distribution is currently infeasible.

To train generalist robots, we must design architectures that lend themselves to generalizing beyond training distributions.
There is a wide variety of work that sets out to do this, but we specifically focus on those methods that do so using 3D representations.
Several lines of work have considered the application of 3D scene representations to end-to-end IL, but these generally are limited to key pose prediction \cite{shridhar2022peract, goyal2023rvt, goyal2024rvt, gervet2023act3d, xian2023chaineddiffuser}, make architectural choices that aren't suitable for out-of-domain transfer \cite{ze2024dp3, ze2024idp3, 3d_diffuser_actor}, or require per-task tuning that is not scalable to multitask settings \cite{wang2024d3fields, wang2023gendp, ze2024dp3}. 

The goal of this paper is to present an architecture explicitly designed for transfer to novel embodiments and camera viewpoints.
We introduce \algabbr{}, an \algname{}. \algabbr{} is an observation backbone that synthesizes outputs from calibrated RGBD cameras into a conditioning vector to be used interchangeably with a variety of IL algorithms when trained end-to-end.
The key insight behind \algabbr{} is to \textit{shift all of the semantic reasoning to a 2D backbone} and use the 3D information \textit{specifically to localize that semantic information with respect to the end-effector}. We do this by aligning feature volumes from a pretrained 2D foundation model with 3D point cloud points and, after carefully post-processing the point cloud, using a learned attention pooling operation to reduce it to a single conditioning vector. 
Using semantic features from a 2D foundation model allows us to bypass the data-intensive process of learning them directly from point clouds, and the post-processing helps the attention pooling step to reason about the points. 

In summary, we contribute the following: 
\begin{itemize}
    \item The \algabbr{} architecture, which extracts conditioning vectors from RGBD inputs to be used interchangeably with a variety of IL algorithms.
    \item Experiments showing that \algabbr{} can combine with a variety of IL algorithms to achieve strong multitask performance and perform high-precision insertions across 93 simulated and 6 real tasks, and maintain this performance with unseen embodiments and camera viewpoints, including a 43.8\% improvement over the next best baseline in our real experiments.
\end{itemize}

\section{Related Work}

\subsection{Imitation Learning for Robotic Manipulation}

Imitation learning (IL) \cite{pomerleau1988alvinn} began with policies mapping observations to actions \cite{pomerleau1988alvinn, zhang2018deep, mandlekar2021learning, florence2019selfsupervised} and has since advanced to energy-based models \cite{florence2022implicit}, diffusion models \cite{diffusionpolicy, 3d_diffuser_actor, ze2024dp3, xian2023chaineddiffuser}, and transformer architectures \cite{act, baku, mete2024quest, vq-bet}, but these methods still suffer outside of their training distributions. Even methods that train on large-scale datasets \cite{rt-1, rt-2, octo, black2024pi0, kim2024openvla, wen2024tinyvla, wen2024diffusionvla, Zawalski24embodiedcot, khazatsky2024droid, rt-x} are brittle to changes in environment such as camera pose or novel embodiments. 
Recent methods have found success transferring to novel embodiments and camera viewpoints through inpainting \cite{chen2024mirage, chen2024roviaug}, but these require expensive inference-time processing or per-embodiment finetuning. 
Rather than attempting to achieve generalization purely through scaling, this paper introduces an architecture which uses 3D representations as a bridge between settings with different embodiments and camera poses.

\subsection{3D Representations in Robot Manipulation}

While the robotics community mostly uses 2D inputs for IL pipelines, a growing body of recent work has achieved success with 3D inputs. PerAct \cite{shridhar2022peract} and RVT \cite{goyal2023rvt, goyal2024rvt} use voxels and simulated camera views respectively to predict end effector keyposes, but this framework assumes tasks that can be neatly decomposed into a sequence of key poses, precluding tasks requiring higher-frequency control or nonlinear trajectories such as folding. Another line of work embeds outputs of foundation models in implicit scene representations \cite{lerftogo2023, yu2024legs, yu2025pogs}, but these require expensive inference-time training, also precluding high-frequency control. 

Most similar to \algabbr{} is a body of work which uses point clouds, a relatively cheap-to-construct scene representation, as input to a policy. \citet{gervet2023act3d} lifts outputs from a frozen pretrained backbone into a point cloud before using this information to predict key poses, and \citet{xian2023chaineddiffuser} builds on this by learning a diffusion policy planner to connect the predicted key poses. 3D Diffuser Actor \cite{3d_diffuser_actor} uses a similar scene representation, self attends between points in the scene and cross attends between scene points, action position samples, language tokens and proprioception. In particular, self-attending between points in the scene requires the agent to reason about its 3D geometry, which is difficult to do in a low-data regime and prone to overfitting. \algabbr{}, on the other hand, is designed to use the 3D information only to localize the semantic information with respect to the end effector, enabling better transfer to novel camera viewpoints. 

The 3D Diffusion Policy (DP3) line of work \cite{ze2024dp3, ze2024idp3} utilizes colorless point clouds as input, followed by a PointNet \cite{qi2016pointnet, qi2017pointnetdeephierarchicalfeature} which pools the cloud into a single vector to be used as conditioning for a diffusion policy. While this design allows generalization to novel object appearances in a single-task setup, its omission of the rich semantic information available to RGB sensors causes it to suffer in multitask or high-precision settings, and its reliance on 3D geometry makes it suffer when evaluated on out-of-distribution camera poses and robot embodiments. Unlike DP3, \algabbr{} uses semantic information from a pre-trained 2D vision backbone to enable multitask IL and alleviate the over-reliance on learning 3D geometric information from scratch.

A final line of work similar to \algabbr{} uses DINO \cite{oquab2023dinov2} features lifted into a 3D representation to achieve impressive instance generalization results \cite{wang2024d3fields, wang2023gendp}, but these works rely on hand-selecting reference features for each task, limiting its scalability to multitask setups or long-horizon tasks with several relevant objects. 
\algabbr{} alleviates this through the attention operation described in Section~\ref{sec:attention}, which automatically learns these features during end-to-end policy learning.
A detailed comparison to related works is available in Appendix~\ref{app:comparison}.

\section{\algabbr{} Architecture}
\label{sec:method}

\begin{figure*}[t!]
    \centering
    \includegraphics[width=\textwidth]{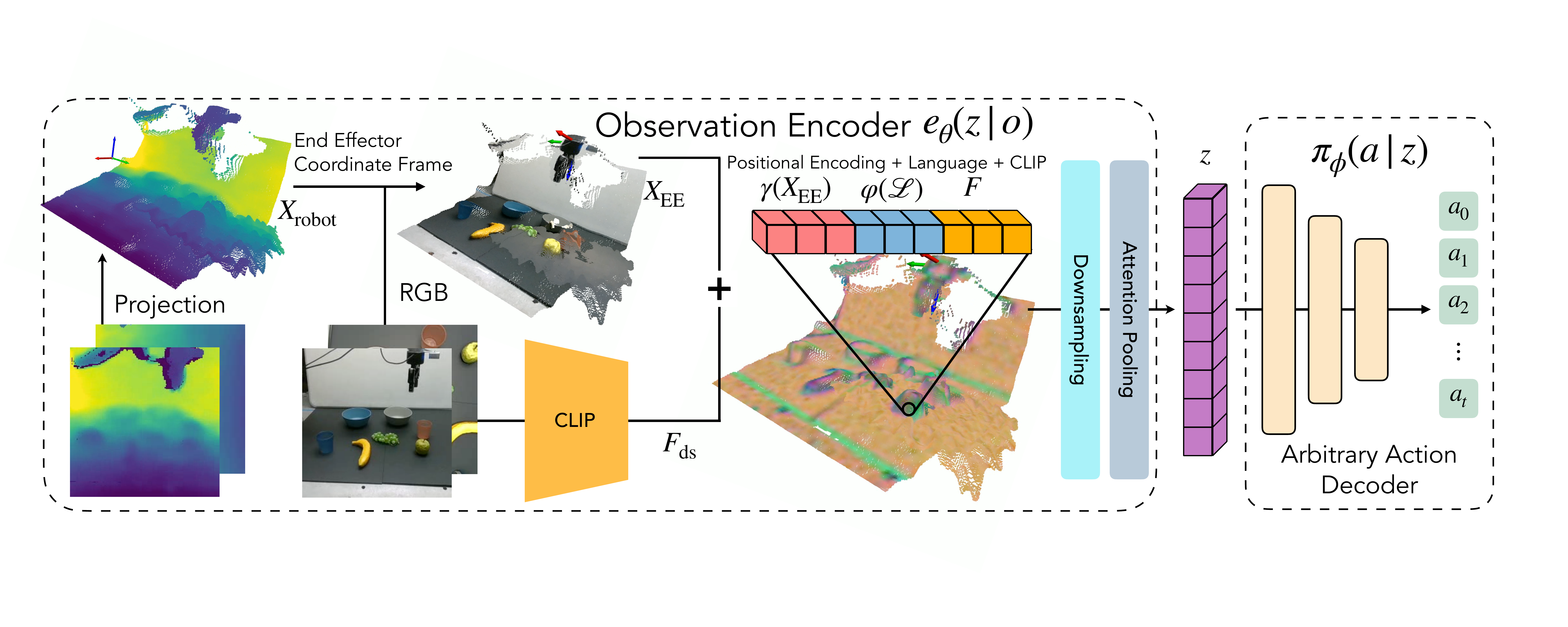}
    \caption{\algabbr{} extracts scene representations from RGBD inputs for use with a variety of imitation learning algorithms. It lifts pre-trained foundation model features into a point cloud, carefully processes that point cloud, and uses attention pooling to compress it into a single vector $z$ to be used as conditioning for end-to-end learning.}
    \label{fig:method}
    
\end{figure*}

Unlike prior methods, which in general make architectural choices leading the policy to infer semantic information from the point cloud, the key idea behind \algabbr{} is to entirely \textit{offload the extraction of semantic information to the 2D backbone}, and instead use \textit{3D information to localize the semantic information} with respect to the end-effector.
\algabbr{} first lifts RGBD images into points paired with semantic and spatial information as described in Section~\ref{sec:3d-rep}. Next, it computes an attention map over points in the cloud in order to reduce the cloud into a single vector as described in Section~\ref{sec:attention}. Finally, \algabbr{} uses that vector as conditioning information for arbitrary downstream action decoders as described in Section~\ref{sec:policy-learning}, and the full pipeline is trained end-to-end, as shown in Figure~\ref{fig:method}.

\textbf{Problem Setup:} We assume access to a dataset $\mathcal{D} = \{\mathcal{L}_i, (o_{i,1}, a_{i,1}), \ldots, (o_{i,T_i}, a_{i,T_i}) \}^N_{i=1}$ consisting of $N$ expert demonstrations with corresponding natural language instructions $\mathcal{L}_i$. Each observation $o_{i,t}=(\mathcal{I}, \mathcal{R}, \mathcal{P})$ is a tuple consisting of a set of RGBD images $\mathcal{I} = \left\{I_1, \ldots, I_{n_{\textrm{cam}}} \in \mathbb{R}^{H \times W \times 4} \right\}$ streaming from $n_{\textrm{cam}}$ cameras. We assume access to a set of corresponding camera calibration matrices mapping 2D image coordinates and depth values to 3D coordinates in the robot's base frame, and proprioceptive information including the end effector pose. \algabbr{}'s goal is to learn an encoder $e_\theta(z|o)$ that subsequently maximizes the success rate of a policy $\pi_\phi(A|z)$ conditioned on $e_\theta$'s outputs when the two are trained together end-to-end.

\subsection{3D Scene Representation}
\label{sec:3d-rep}

We construct a 3D scene representation with the necessary information and structure for our end goal of compressing it into a single compact conditioning vector. Similarly to \cite{3d_diffuser_actor, gervet2023act3d, xian2023chaineddiffuser, wang2023gendp, wang2024d3fields}, we use camera calibration data to lift semantic features from several viewpoints into a single fused point cloud. This allows \algabbr{} to reason about the scene without extracting semantic information from the point cloud, which is prone to overfitting on smaller datasets.

Specifically, for each image $I_i$, we extract an intermediate feature volume using a pretrained CLIP ResNet \cite{clip} and project it to feature $F_i \in \mathbb{R}^{h \times w \times d}$, where $h < H$ and $w < W$. We deproject each image into a point cloud, transform it into the robot base frame, and resize it to $X_i \in \mathbb{R}^{h \times w \times 3}$ using nearest neighbors interpolation. Finally, we flatten and concatenate these to construct features $F \in \mathbb{R}^{m \times d}$ and point locations $X_{\textrm{robot}} \in \mathbb{R}^{m \times 3}$ with $m = n_{\textrm{cam}}hw$.

\textbf{End Effector Coordinate Frames:} 
Wrist camera observations are often used to help robots reason about the relative positions of the end effectors and target objects \cite{mandlekar2021matters}.
Applying the same intuition to point clouds, we use proprioceptive information to transform them into the end effector's coordinate frame, giving us the new point cloud $X_{\textrm{EE}}$. 
\citet{liu2022frame} notes that this change can lead to improved data efficiency, but we observe that it is especially helpful for cross-embodiment learning, where it is important to reason about the relative positions of the end effector and the scene without overfitting to the training distribution embodiment.

\textbf{Positional Encoding:} As described in \cite{rahaman19spectral, mildenhall2020nerf}, neural networks are not well suited to fitting data with high-frequency variation. In robotics, where a small difference in relative position between the end effector and object can determine the success of a rollout, this issue is especially important. Thus, as in \cite{mildenhall2020nerf} we encode each 3D point using a Fourier basis of increasing frequencies, mapping raw positions into a higher-dimensional space that is more amenable to learning complex functions.
We apply 
    $\gamma(x) = \big( \sin{(2^0\pi x)},  \cos{(2^0\pi x)}, \ldots, \sin{(2^{L-1}\pi x)},  \cos{(2^{L-1}\pi x)}\big)$
with $L=10$, which lifts the positions into $\hat{x} \in \mathbb{R}^{60}$. Let $\hat{X}_{\textrm{EE}} = \gamma(X_{\textrm{EE}})$ denote the lifted point cloud.

\subsection{Point Cloud Reduction}
\label{sec:attention}

After constructing the 3D scene representation as described in Section~\ref{sec:3d-rep}, we reduce it into a vector to be used as conditioning for the downstream policy. Our method is designed to do so without extracting semantic information from the relative locations of the points, instead using the information from the 2D backbone, and using 3D only to localize this information with respect to the robot.
The key steps in this process are cropping, downsampling, and the final learned reduction.

\textbf{Cropping:} 
Rather than carefully selecting tight bounds \cite{ze2024dp3,wang2023d} or forgoing cropping altogether \cite{3d_diffuser_actor, ze2024idp3}, we take an intermediate stance, noting that especially with novel camera poses some cropping can help to ignore OOD background points.
Specifically, we crop scenes such that the entire table is visible but objects outside the table are not. Next, since the positive $z$ axis points outward from the end-effector in $X_{\textrm{EE}}$, we filter points with negative $z$ values, cropping out most of the arm and aiding cross-embodiment transfer. More details and a sensitivity study are in Appendix~\ref{app:sim-results}.

\textbf{Downsampling:}
When changing camera poses the distribution of points on task-relevant and task-irrelevant (eg. the table) objects changes drastically, and we address this through a strategic choice in downsampling algorithm.
Unlike \cite{ze2024dp3, ze2024idp3}, which performs furthest point sampling based on Cartesian coordinates, we adopt the approach of \citet{3d_diffuser_actor} and downsample according to $\ell_2$ distance between the features $F$.
We find this strategy to yield point clouds that are more concentrated around the task-relevant objects and more robust to changes in camera pose, providing visualizations of this in Appendix~\ref{app:sim-results}. We downsample to $\hat{X}_{\textrm{EE},\textrm{ds}} \in \mathbb{R}^{p \times 60}$ and $F_{\textrm{ds}} \in \mathbb{R}^{p \times d}$ where $p < m$.

\textbf{Language Embeddings:}
Prior work has found FiLM \cite{film} modulation based on language inputs in the RGB backbone to help policies reason about their tasks in multitask settings \cite{liu2024libero, baku}. Likewise, we concatenate projected CLIP language embeddings $\varphi(\mathcal{L}) \in \mathbb{R}^d$ to the representations of the points in our point cloud. 
We define our final point cloud $P \in \mathbb{R}^{p \times (2d + 60)}$ to be the concatenation of positional encodings $\hat{X}_{\textrm{EE},\textrm{ds}}$, feature vectors $F_{\textrm{ds}}$ and language embeddings $\varphi(\mathcal{L})$.

\textbf{Encoding Extraction:}
After post processing our point cloud, we must reduce it into a conditioning vector without risk of overfitting to specific attributes of the training scenes. We learn an attention map over the points and use the resulting attention pooling operation to output a final vector $z$ which we use as conditioning for the final policy. Concretely we learn key and value MLPs $K_\theta: \mathbb{R}^{2d + 60} \rightarrow \mathbb{R}^{d_k}$ and $V_\theta: \mathbb{R}^{2d + 60} \rightarrow \mathbb{R}^{d_e}$, and learn a fixed query $q \in \mathbb{R}^{d_k}$. The final embedding $z$ is defined to be $z = \mathrm{softmax}\left( \frac{q K_\theta(P)}{\sqrt{d_k}} \right) V_\theta(P)$. We find that in settings with very large changes in camera pose, this operation doesn't attned to out-of-distribution points, which helps with generalization to the new viewpoint. Visualizations of these attention maps are provided in the supplementary material.

\subsection{Policy Learning}
\label{sec:policy-learning}

After extracting the encoding vector $z$, we use it to train a policy. \algabbr{} is designed as a modular encoder compatible with a range of downstream action decoders, and we combine it with Action Chunking Transformer (ACT) \cite{act}, Diffusion Policy (DP) \cite{diffusionpolicy}, and BAKU \cite{baku}.
For each method we use \algabbr{} to extract $z$ from the observations and use it as conditioning to the policy. This process is described for each policy choice in greater detail in Appendix~\ref{app:method-detail}.

\section{Simulated Experiments}
\label{sec:experiments}

In this section, we set out to answer the following questions:
(1) Does \algabbr{} have the capacity to achieve strong performance in multitask and high-precision settings?
(2) How does \algabbr{} perform in zero-shot cross-embodiment settings compared with existing encoders?
(3) How does \algabbr{} perform in zero-shot camera viewpoint change experiments compared with existing encoders?
(4) How do the various design decisions described in Section~\ref{sec:method} affect \algabbr{}'s performance?

\label{sec:benchmarks}
\paragraph{Benchmarks}
\textbf{LIBERO-90} \cite{liu2024libero} is a multitask learning benchmark consisting of 90 rigid-body and articulated manipulation tasks which we use to study \algabbr{}'s multitask reasoning capabilities. \textbf{MimicGen} \cite{mandlekar2023mimicgen} includes 3 difficult insertion tasks, and we use it to study \algabbr{}'s suitability for settings requiring precise control. Further details about both benchmarks are available in the Appendix~\ref{sec:exp-details}.

\begin{figure}
    \centering
    \includegraphics[width=\textwidth]{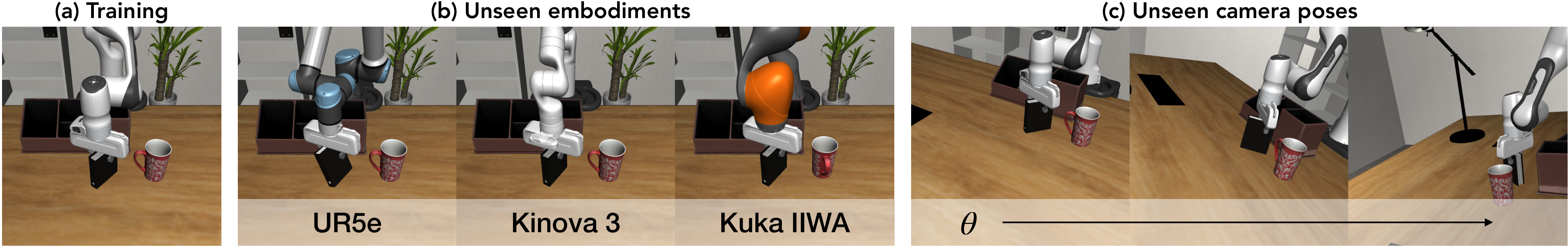}
    \caption{We train on the Franka Panda and viewpoint shown in (a). Then, we evaluate zero-shot with the UR5e, Kinova3 and IIWA (b) embodiments, and unseen camera poses (c). 
    } 
    \label{fig:domain-changes}
\end{figure}

\subsection{Baselines}
\label{sec:baselines}

The goal of these experiments is to understand how a variety of observation backbones affect the properties of the final learned policy. To this end, for each of the action decoders mentioned in Section~\ref{sec:policy-learning}, our baselines run the same algorithm while replacing the observation encoder $e_\theta$ with the following backbones: \textbf{(1) RGB}: The CNN backbone of a ResNet-18 \cite{resnet} with pre-trained ImageNet weights \cite{deng2009imagenet} followed by a spatial softmax \cite{levine2016endtoend}, fine-tuned. 
\textbf{(2) RGBD}: Similar to RGB, but with an added depth channel.
\textbf{(3) DP3}: The encoder from 3D Diffusion Policy \cite{ze2024dp3}, which lifts the depth information into a colorless point cloud and processes that cloud into a conditioning vector using PointNet \cite{qi2016pointnet}.
Finally, we compare to \textbf{3D Diffuser Actor (3DDA)} \cite{3d_diffuser_actor}, which lifts a semantic feature map into the point cloud and denoises action sequences (similarly with DP) by self-attending on the set of points and cross-attending between points and noisy action sequences. 
DP3 and 3DDA have access to the same world frame cropping as \algabbr{}.

All baselines have access to the same set of camera inputs, camera calibration data, proprioceptive information and language instructions as \algabbr{}. We train 5 seeds for each algorithm variation and perform 10 evaluation rollouts per environment per seed for LIBERO and 100 rollouts for MimicGen. We present aggregate results here and full numerical results in Appendix~\ref{app:sim-results}.

\subsection{In-Distribution Behavior Cloning}

\begin{wraptable}{r}{0.5\textwidth}
    \centering
    \vspace{-0.2in}
    \caption{\textbf{In-distribution Success Rate.} LIBERO-90 results are aggregated across ACT, BAKU and DP, and MG results are from DP.}
    \label{tab:mt}
    \resizebox{1.0\linewidth}{!} {
    \begin{tabular}{l|ccccc}
    \toprule
         & RGB & RGBD & DP3 & 3DDA & \algabbr{} \\
         \midrule
       LIBERO-90  & $\mathbf{90.9}$ & $78.8$ & $70.7$ & $83.7$ & $\mathbf{90.0}$ \\
       \rowcolor[HTML]{EFEFEF}
        MG-Coffee & $\mathbf{74.4}$ & $19.8$ & $1.8$ & $46.2$ & $\mathit{70.8}$ \\        
        MG-Square & $60.8$ & $45.2$ & $1.6$ & $\mathbf{66.2}$ & $\mathit{61.6}$ \\
        \rowcolor[HTML]{EFEFEF}
        MG-Threading & $\mathit{42.4}$ & $0.2$ & $0.2$ & $24.8$ & $\mathbf{44.0}$ \\
    \bottomrule
    \end{tabular}
    }
    \vspace{-0.2in}
\end{wraptable}

In this experiment, we set out to understand whether \algabbr{} has the modeling capacity to achieve strong success rates in multitask settings and high-precision insertion tasks. 

We present results in Table~\ref{tab:mt}, where LIBERO-90 results are aggregated across all action decoders and MimicGen results use DP. We see that \algabbr{} does in fact have the modeling capacity to achieve strong results on in-distribution settings, achieving similar performance to baselines in all experiments, and achieving the strongest overall success rate for the MimicGen Threading task.

One takeaway is that \algabbr{} has far stronger performance than DP3, which does not include semantic features, indicating that \textit{sparse point clouds without semantic features are not enough to succeed in multitask or high-precision settings}. Another surprising takeaway is that RGBD performs worse than RGB despite having access to more information. This is likely because the wrist camera helps with the spatial reasoning that depth would otherwise help with and the depth is simply extra information that the policy overfits to.

\begin{figure}[t]
    \centering
    \includegraphics[width=0.35\textwidth]{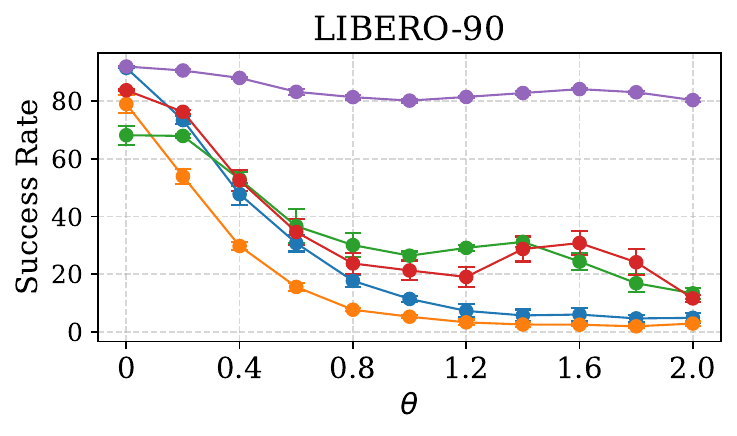}
    \includegraphics[width=0.21\textwidth]{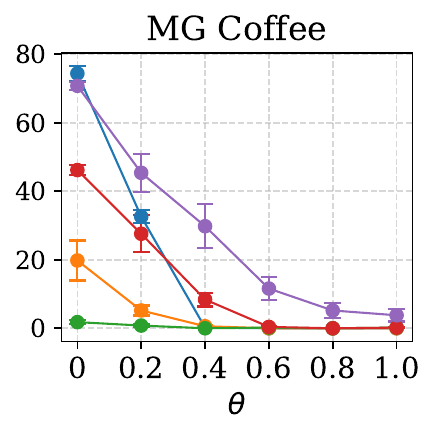}
    \includegraphics[width=0.21\textwidth]{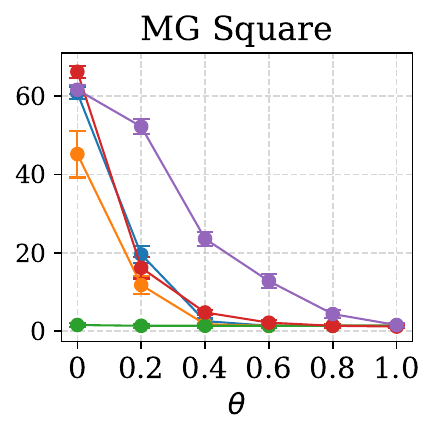}
    \includegraphics[width=0.21\textwidth]{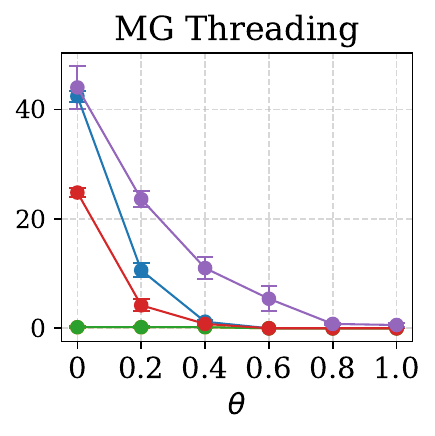}
    \vspace{-5pt}
    \includegraphics[width=0.6\textwidth]{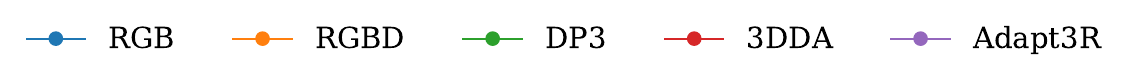}
    \caption{\textbf{Unseen Camera Pose.} We rotate the scene camera by $\theta$ radians about the vertical axis through the end-effector starting position. LIBERO-90 results use BAKU and MimicGen (MG) results use DP.}
    \label{fig:cam-change}
    \vspace{-5pt}
\end{figure}

\subsection{Unseen Camera Poses}
\label{sec:unseen-cam}

In this section, we investigate whether \algabbr{} leads to improved performance when rolling out with unseen camera poses. We train using the original viewpoints from the dataset, which include a scene camera and a wrist camera. At inference time, we rotate the scene camera about the vertical axis through the end-effector starting position as shown in Figure~\ref{fig:domain-changes}.
Results are presented in Figure~\ref{fig:cam-change}, where LIBERO-90 results use BAKU and MimicGen results use DP. We see that \algabbr{} consistently gives a substantial boost in performance over comparison observation encoders, especially in settings with very large changes in camera viewpoint, indicating that our scene representation does a good job facilitating this type of generalization. Notably, \algabbr{} maintains $>80\%$ success rate on LIBERO-90, and has the only nonzero MimicGen success rates after $\geq 0.6$ radians of rotation. 

\begin{figure}[!t]
    \centering
    \vspace{-.2in}
    
    \includegraphics[width=0.23\textwidth]{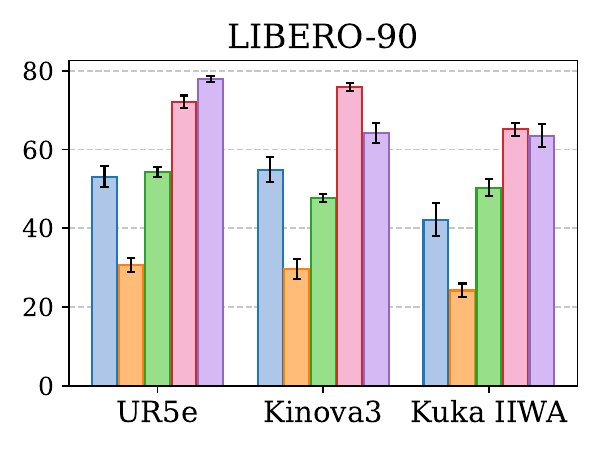}
    \includegraphics[width=0.23\textwidth]{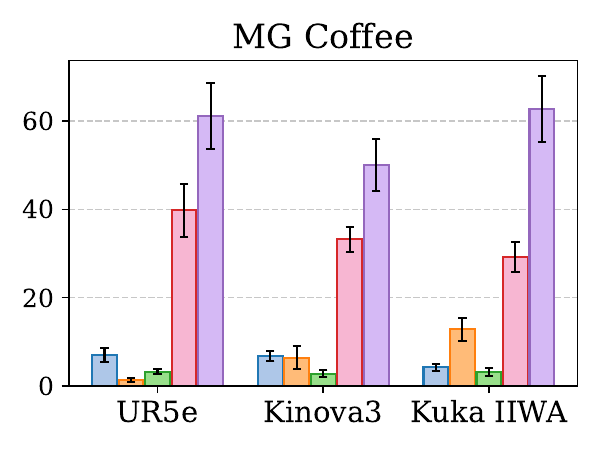}
    \includegraphics[width=0.23\textwidth]{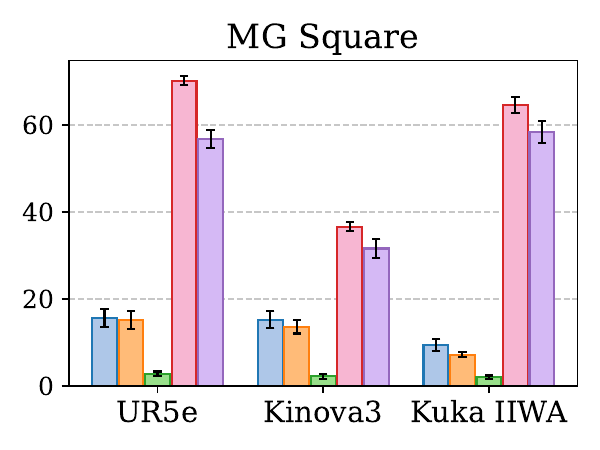}
    \includegraphics[width=0.23\textwidth]{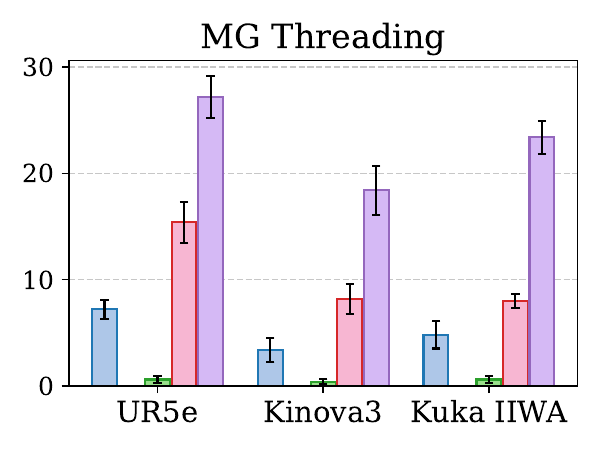}
    \includegraphics[width=0.6\textwidth]{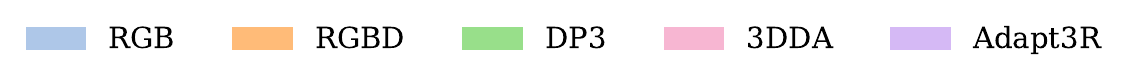}
    \caption{\textbf{Cross Embodiment.} We evaluate zero-shot with three unseen embodiments. LIBERO-90 results aggregate across all action decoders and MimicGen (MG) results use DP. \algabbr{} and 3DDA consistently outperform comparisons, indicating that semantic-aligned point clouds are conducive to embodiment transfer.
    }
    \label{fig:xe-results}
    \vspace{-5pt}
\end{figure}

DP3 and 3DDA do not do well in this setting. Both of them use calibrated depth cameras to create a point cloud which should intuitively be viewpoint-agnostic. We speculate that this is because the changes in camera position change the distribution of points in the scene and these methods, which condition more on the 3D geometry of the scene, are more brittle to this change.

\subsection{Cross Embodiment}

In this section, we investigate whether \algabbr{} leads to improved performance in a zero-shot cross-embodiment setting, training on the Panda robot and evaluating on the embodiments in Figure~\ref{fig:domain-changes}.
Results are shown in Figure~\ref{fig:xe-results}, where LIBERO-90 is aggregated across all action decoders and MimicGen uses DP. We see that \algabbr{} consistently outperforms RGB, RGBD and DP3, especially in the MimicGen experiments where reasoning about scene camera inputs is more important, indicating its superiority in this setting among standalone observation encoders. \algabbr{} achieves similar performance to 3DDA in LIBERO and MG Square and superior performance for Coffee and Threading, indicating that point clouds with semantic features are consistently an effective choice for cross-embodiment transfer.

\subsection{Inference Speed}

\begin{wraptable}{r}{0.4\textwidth}
    \centering
    \vspace{-.2in}
    \caption{Inference rates (Hz) for \algabbr{} and baselines.}
    \label{tab:speed}
    \resizebox{\linewidth}{!} {
    \begin{tabular}{cccc}
    \toprule
         RGB & DP3 & 3DDA & \algabbr{} \\
         \midrule
         132.2 & 146.1 & 2.6 & 44.1 \\
    \bottomrule
    \end{tabular}
    }
\end{wraptable}

In Table~\ref{tab:speed} we present inference speeds for \algabbr{} and the main comparisons. We observe that while RGB and DP3 run at high speed, they perform poorly, and in contrast 3DDA performs well but at an untenable 2.6~\unit{Hz}. \algabbr{} provides high performance at a 44.1~\unit{Hz}, which is faster than the input rate of prevalent RGBD cameras ($\sim$30~\unit{Hz}).

\subsection{Ablation Studies}
In this section, we ablate several design decisions that contribute to \algabbr{}, hoping to understand how these decisions affect its multitask modeling capacity and ability to generalize to novel settings. In addition, we experiment with backbone choice, world frame cropping choice and pooling architecture in the supplement. Table~\ref{table:ablations} presents ablated versions of \algabbr{} when trained together with BAKU.

\begin{table}[!t]
\centering
\begin{minipage}{0.29\textwidth}
\centering
\caption{\textbf{Ablations.} In-distribution results, cross embodiment results and camera change results for ablated versions of \algabbr{}. We find that the design decisions lead to substantial improvements in generalizability.}
\vspace{-0.05in}
\label{table:ablations}
\end{minipage}
\begin{minipage}{0.7\textwidth}
\centering
\vspace{0.05in}
\resizebox{\linewidth}{!} {
    \begin{tabular}{l|c|ccc|ccc|c}
    \toprule
    \rowcolor[HTML]{FFDEB4}
    Variant & Orig. & UR5e & Kinova3 & IIWA & $\theta=0.4$ & $\theta=1.0$ & $\theta=2.0$ & Average \\
    \midrule
    No EECF & $83.0$ & $69.3$ & $61.6$ & $61.2$ & $81.6$ & $79.3$ & $78.4$ & $73.5$ \\
    \rowcolor[HTML]{EFEFEF}
    No EE Crop & $91.3$ & $77.8$ & $67.4$ & $70.0$ & $83.4$ & $82.2$ & $82.1$ & $79.2$ \\
    No Image Features & $60.9$ & $48.7$ & $37.2$ & $43.5$ & $2.9$ & $1.9$ & $4.4$ & $28.5$ \\
    \rowcolor[HTML]{EFEFEF}
    RGB Point Cloud & $60.5$ & $48.2$ & $35.7$ & $42.5$ & $3.7$ & $2.1$ & $4.8$ & $28.2$ \\
    No Lang. Features & $91.3$ & $78.7$ & $62.7$ & $64.9$ & $\mathbf{84.7}$ & $82.2$ & $82.4$ & $78.1$ \\
    \rowcolor[HTML]{EFEFEF}
    No Positional Encoding & $84.3$ & $69.9$ & $61.3$ & $66.9$ & $58.0$ & $70.5$ & $70.9$ & $68.8$ \\
    Position-Based FPS & $91.6$ & $76.8$ & $59.6$ & $64.2$ & $71.7$ & $65.1$ & $70.9$ & $71.4$ \\
    \rowcolor[HTML]{EFEFEF}
    No Attention & $\mathbf{92.0}$ & $\mathbf{82.7}$ & $\mathbf{70.0}$ & $69.5$ & $71.3$ & $64.1$ & $64.9$ & $73.5$ \\
    Ours & $\mathbf{91.9}$ & $80.1$ & $65.3$ & $\mathbf{71.8}$ & $\mathbf{84.7}$ & $\mathbf{82.9}$ & $\mathbf{82.9}$ & $\mathbf{80.0}$ \\
    \bottomrule
    \end{tabular}
}
\end{minipage}
\end{table}

\textbf{No EECF} and \textbf{No EE Crop} removes the transformation to the end effector's coordinate frame and cropping in that frame respectively. Results show that both of these lead to improvements across the board as they help the robot reason about the end effector's position relative to objects in the scene.

\textbf{No Image Features} and \textbf{RGB Point Cloud} remove image features $F$ and replace them with nothing or the original RGB values respectively. Overall we see that removing this information leads to a substantial drop in performance, especially when generalizing to new camera poses. 

\textbf{No Lang Features} does not concatenate the language embeddings $\varphi(\mathcal{L})$. We see that adding them leads to a modest improvement across the board.

\textbf{No Positional Encoding} does not lift the $x \in \mathbb{R}^3$ to $\gamma(x) \in \mathbb{R}^{60}$. We find this change to be important, especially when generalizing to new viewswhere it leads to an average $11.2\%$ improvement.

\textbf{Position-Based FPS} downsamples point clouds according to Cartesian coordinates rather than feature vectors. This design choice leads to a respectable benefit across the board, improving success rates by an average of $6.6\%$. It is especially helpful when transferring to novel camera views.

\textbf{No Attention} uses max pooling instead of attention pooling. Attention pooling is particularly helpful when transferring to new viewpoints, where it gives less attention to out-of-distribution points.

\section{Real Experiments}
\label{sec:real-exps}

\begin{figure*}
    \centering
    \begin{subfigure}[b]{0.3\textwidth}
        \centering
        \includegraphics[width=\textwidth]{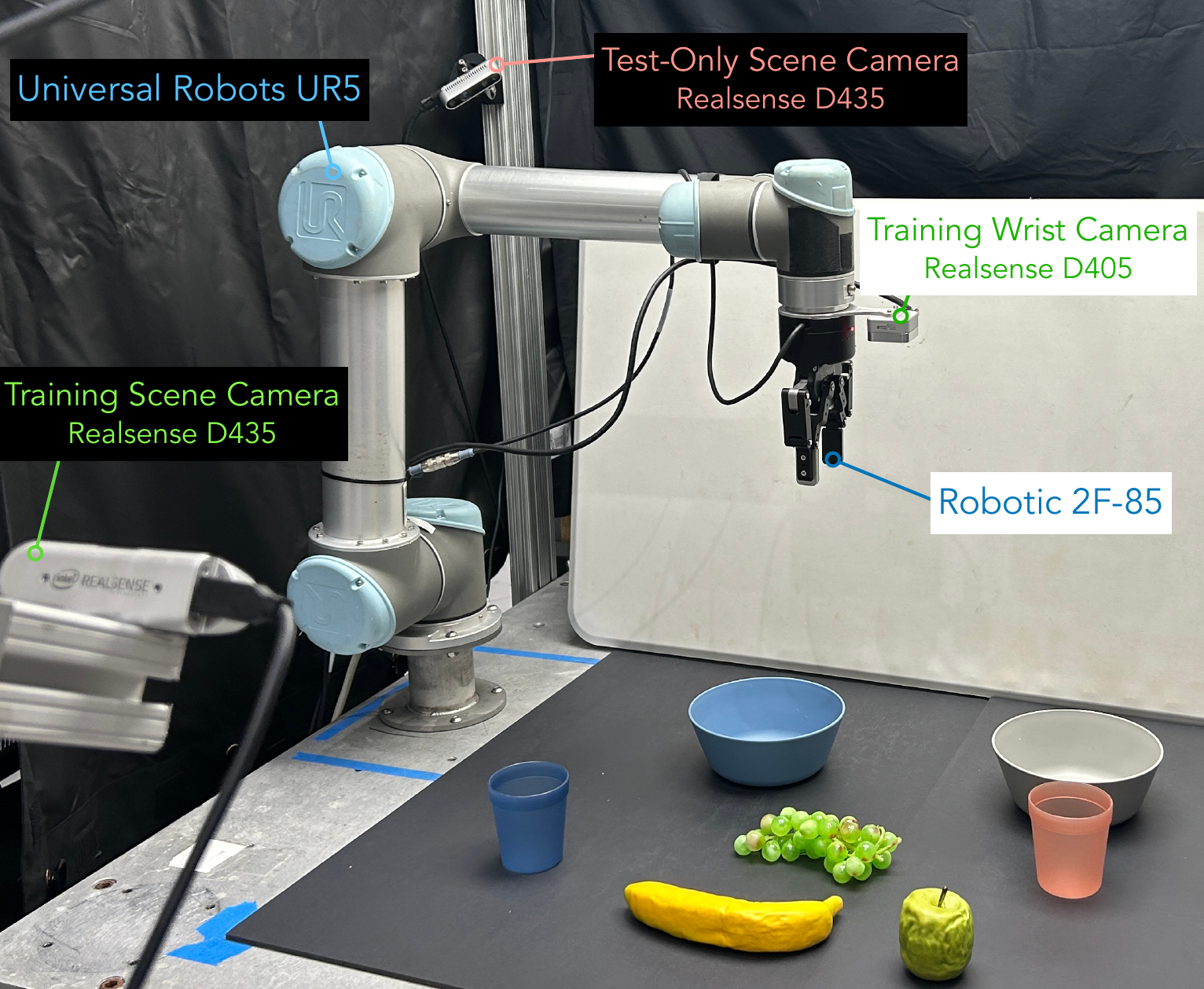}
        \captionsetup{font=scriptsize}
        \caption{Real experiment setup}
        \label{fig:real-diagram}
    \end{subfigure}
    \begin{subfigure}[b]{0.32\textwidth}
        \centering
        \includegraphics[width=\textwidth]{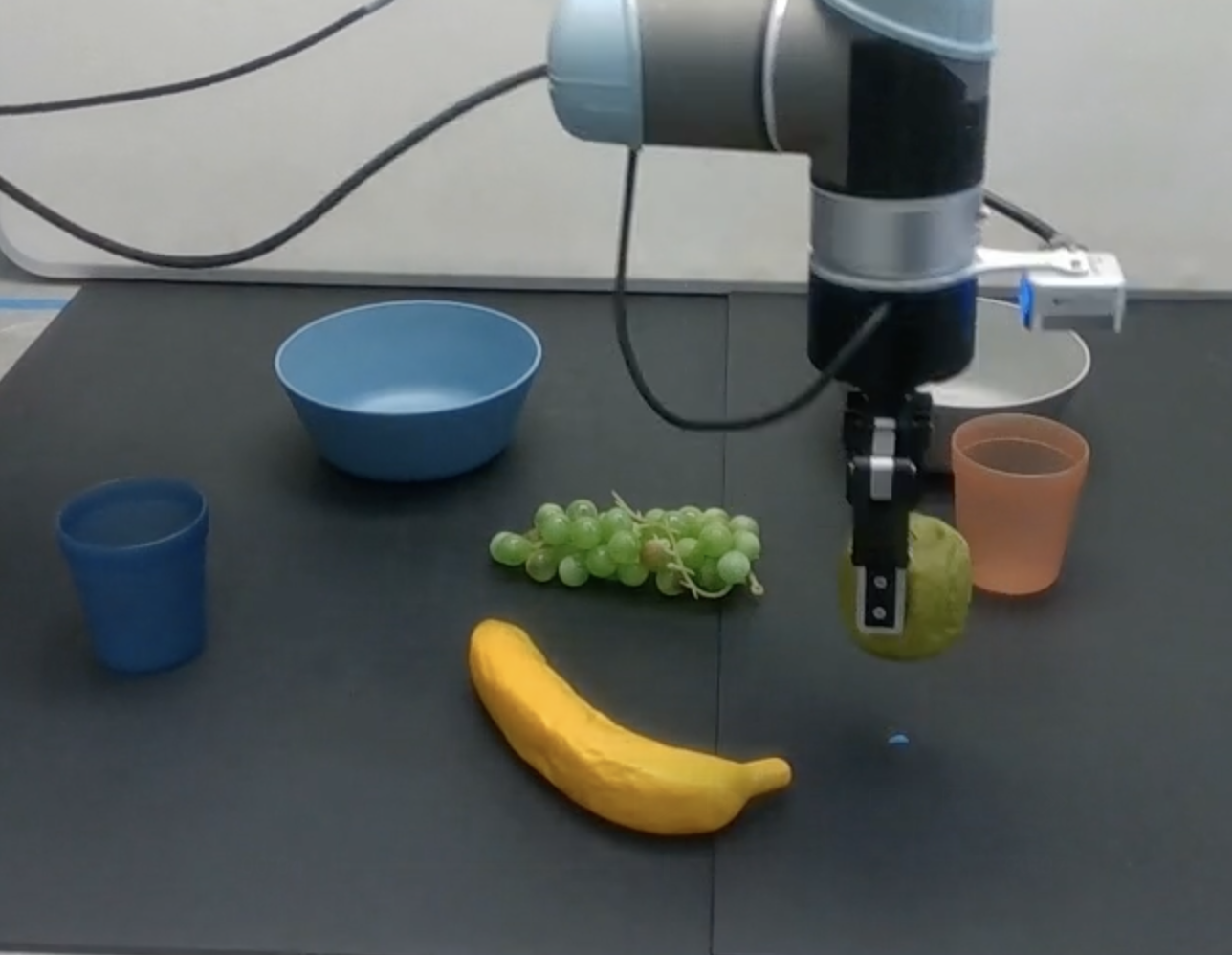}
        \captionsetup{font=scriptsize}
        \caption{Training Viewpoint}
        \label{fig:real-train-view}
    \end{subfigure}
    \begin{subfigure}[b]{0.32\textwidth}
        \centering
        \includegraphics[width=\textwidth]{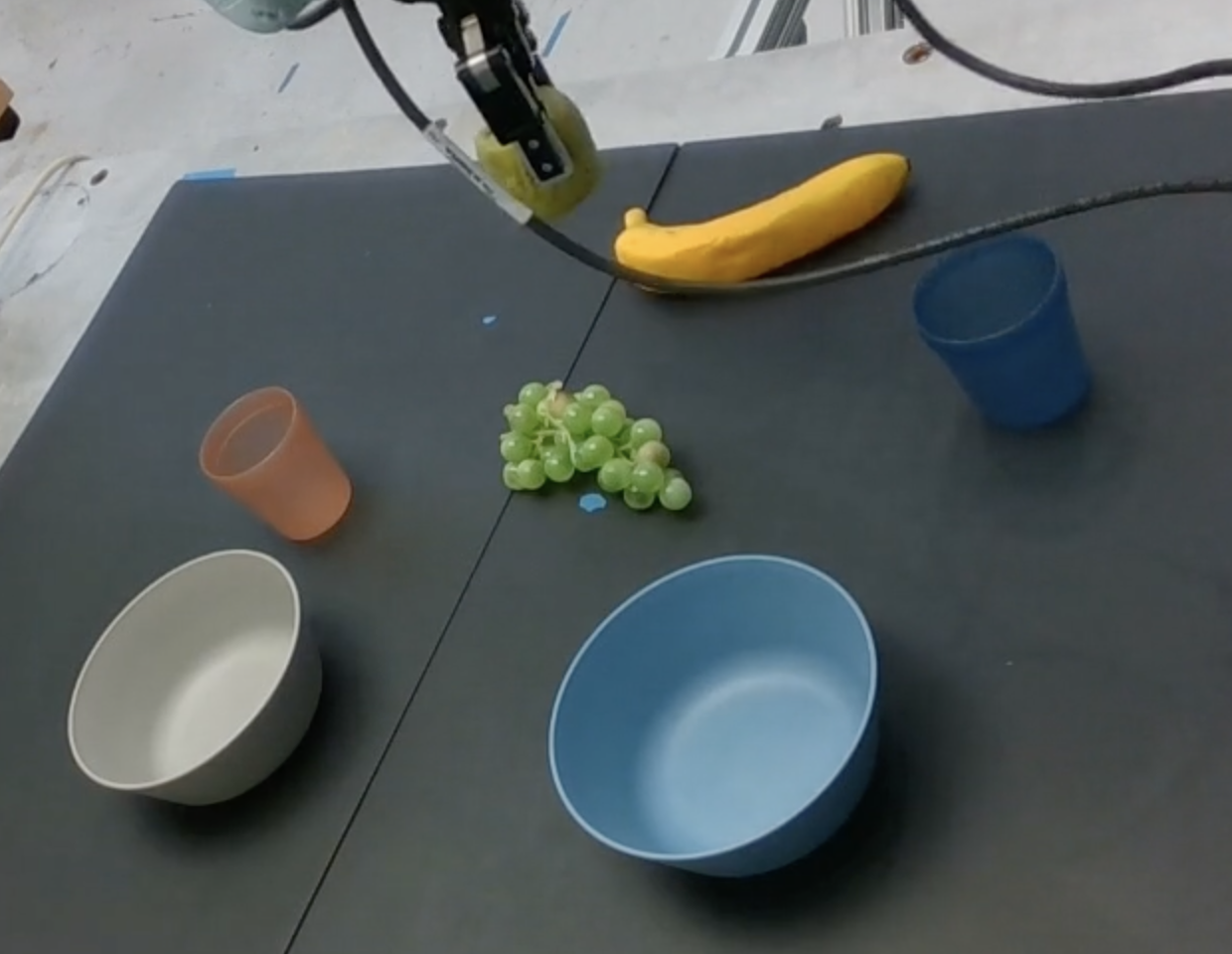}
        \captionsetup{font=scriptsize}
        \caption{Zero-Shot Evaluation Viewpoint}
        \label{fig:real-test-view}
    \end{subfigure}
    
    \caption{\textbf{Real-Robot Setup}. (a) Illustration of Hardware. (b) The viewpoint used to train all policies. (c) The viewpoint used for our zero-shot evaluation experiments.}
    \label{fig:real_setup}
\end{figure*}

In this section we investigate \algabbr{}'s applicability to a real-world multitask IL benchmark with 6 tasks, shown in Figure~\ref{fig:real_setup}, and test whether it shows the same performance with unseen camera poses.

\label{sec:real-setup}
\paragraph{Experiment Setup}
For this experiment we compare the \algabbr{} backbone to RGB, DP3 and 3D Diffuser Actor (3DDA). We train with data from the scene camera shown in Figure~\ref{fig:real-train-view} and, for our zero-shot experiments, evaluate with the scene camera shown in Figure~\ref{fig:real-test-view}. 
For our results we record \textit{progress towards task completion} as opposed to raw success rates. Extended details are in Appendix~\ref{app:real-details}.

\subsection{Results}
\begin{wrapfigure}{r}{0.5\textwidth}
    \centering
    \vspace{-0.2in}
    \includegraphics[width=\linewidth]{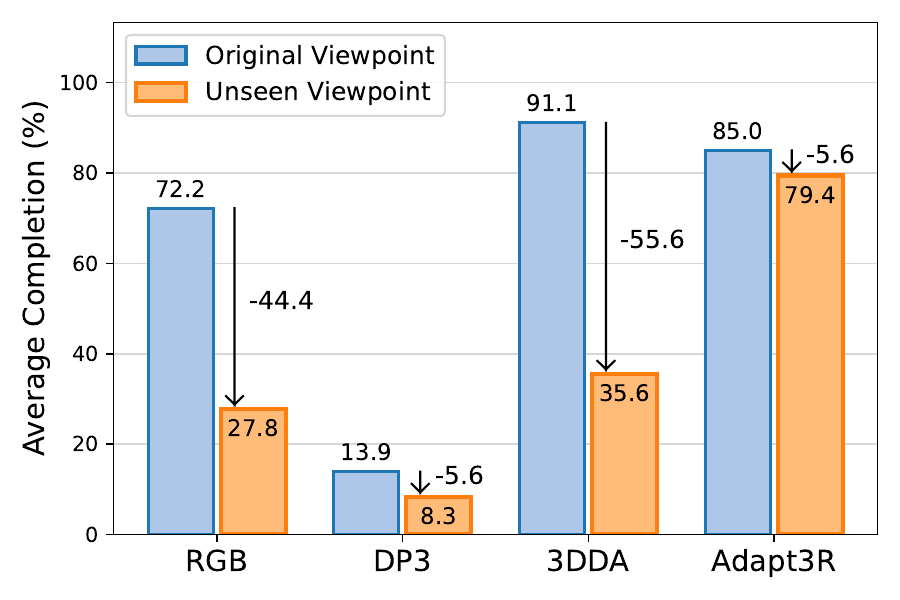}
    \caption{\textbf{Real Benchmark Results}. We see that \algabbr{} achieves strong in-distribution performance and retains performance under the change in viewpoint, while baselines do not.}
    \label{fig:real-res}
    \vspace{-0.2in}
\end{wrapfigure}

We present results from the real benchmark in Figure~\ref{fig:real-res}. 
\algabbr{}'s in-distribution performance is similar to 3DDA and shows an improvement over the RGB and DP3 baselines. DP3's performance was particularly poor, likely because it omits RGB semantic information and uses point locations alone to infer information about the scene which is unreliable with noisy depth estimates from the real sensor. This indicates the importance of semantic features for real-world IL from 3D inputs.

Next, we evaluate policies after replacing the scene camera input with the unseen camera viewpoint shown in Figure~\ref{fig:real-test-view}. We see that while RGB and 3DDA drop in performance by 44.4\% and 55.6\% respectively,
\algabbr{}'s performance drops by less than 6\%, indicating strong performance with the unseen camera pose. 
Notably, it achieves a higher average task completion out-of-distribution than RGB and DP3 do in-distribution. 
Although 3DDA achieves strong in-distribution performance it suffers under changes in camera pose, indicating that its architecture overfits to the training distribution. On the other hand, \algabbr{}, whose architecture is explicitly designed to encourage the learner not to excessively condition on the relative locations of the points, does not have this issue and reliably completes the task with the new camera pose.

\section{Conclusion}

In this paper, we present \algname{} (\algabbr{}), a general-purpose observation encoder using 3D scene representations. The key idea behind \algabbr{} is to offload the inference of semantic information to a pretrained 2D vision foundation model, and use 3D information specifically to localize those semantic features with respect to the end-effector pose. 
Experimental results demonstrate that \algabbr{} facilitates strong performance in multitask BC setups and high-precision insertion tasks, and leads to substantial improvements for zero-shot transfer to novel viewpoints and embodiments. Experiments in a real-world multitask IL benchmark show that \algabbr{} is deployable on a physical robot and lends itself equally well to domain transfer. We hope this work can be a stepping stone towards the application of 3D representations for general autonomy.

\newpage
\section{Limitations} 
The most important limitation of this work is its reliance on both depth information and camera calibration matrices to facilitate the creation of fused multi-view point clouds, and the transformation to the end-effector's coordinate frame. 
Collecting this information can be cumbersome, and it is sometimes missing or inaccurate in large-scale datasets \cite{khazatsky2024droid, vuong2023open, ebert2021bridge, walke2023bridgedata}. Furthermore, depth cameras fail on objects that are particularly reflective or translucent. We are hopeful that in the future these issues can be resolved by methods using foundation models to extract calibration, like \citet{khazatsky2024droid} did using DUSt3R \cite{wang2024dust3r}, and with more accurate learned depth estimators \cite{wen2025foundationstereo}.

Another limitation is that the transformation to the end effector's coordinate frame raises questions about the applicability of \algabbr{} to bimanual settings. A simple way to handle this would be to simply condition on two \algabbr{} vectors, one for each hand or convert to a robot torso frame. In the future we hope to explore 3D representations with less restrictive choices of canonical coordinate frame.

A third limitation of the work is its rather narrow treatment of ``cross embodiment''. Specifically, we only consider cross embodiment learning between robots with shared action spaces (6D pose prediction), which allows zero-shot transfer to unseen embodiments. In this setting cross embodiment learning is merely a perception problem, but it is unrealistic to assume a shared action space across a large diverse dataset. In the future we'd like to train more complex architectures, perhaps using a separate head for each action space as in \cite{nvidia2025gr00tn1openfoundation}. With this being said, we hope this paper's core claim--that \algabbr{} helps to close the perception gap between embodiments--is useful for more general cross-embodiment settings.

\section*{Acknowledgments}

We would like to thank Akshay Krishnan and Jeremy Collins for helpful discussions and for help with writing the paper.

\bibliography{main}

\clearpage
\newpage
\onecolumn

\begin{appendix}

\section{Comparison to Similar Methods}
\label{app:comparison}
\begin{figure*}[!h]
    \centering
    \begin{subfigure}[b]{\textwidth}
        \centering
        \includegraphics[width=\textwidth]{figures/fig2_method.pdf}
        
        \caption{\algabbr{}}
        \label{fig:method-comp-adapt3r}
    \end{subfigure}

    \begin{subfigure}[b]{0.37\textwidth}
        \centering
        \includegraphics[width=\textwidth]{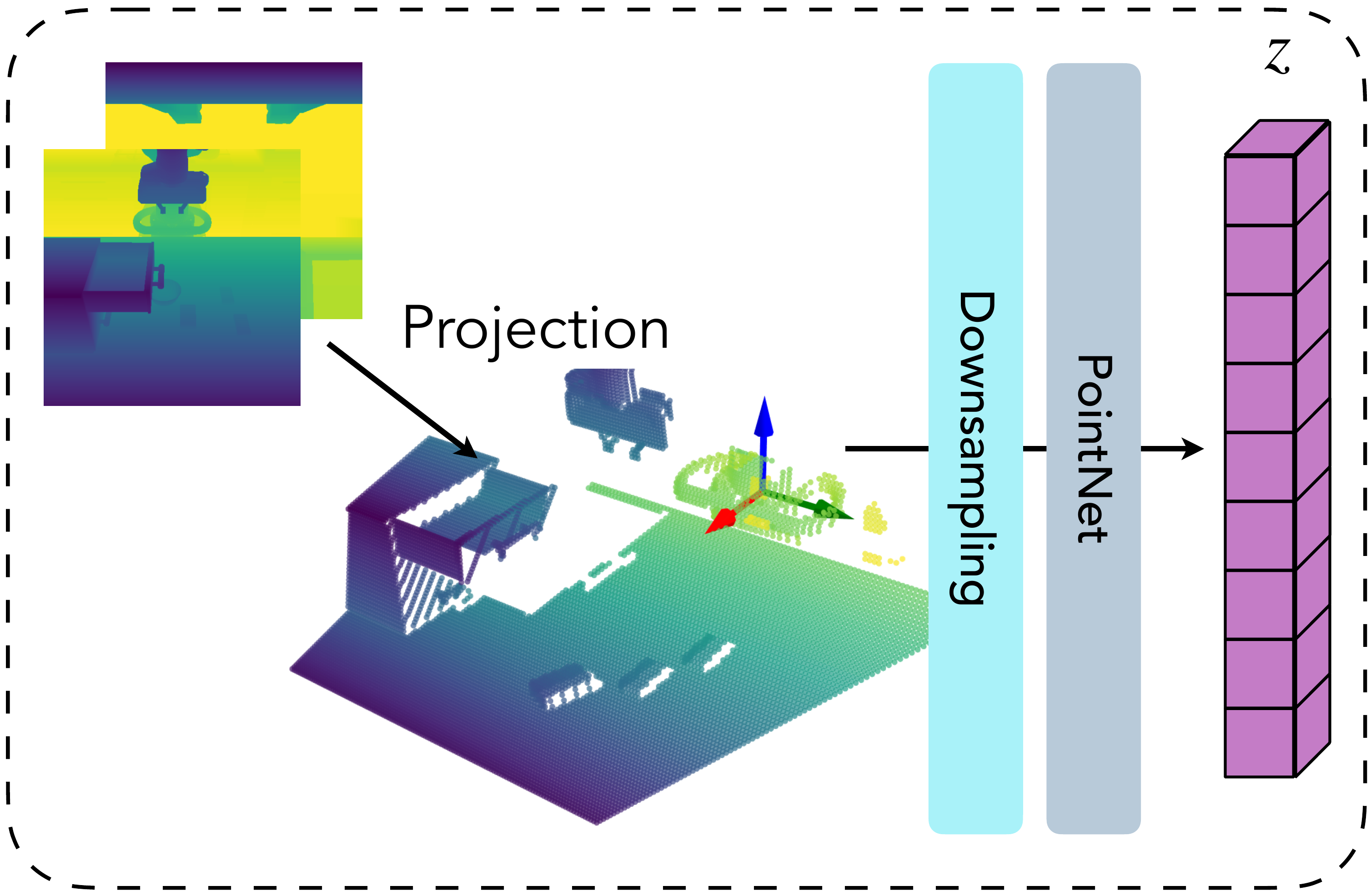}
        \caption{DP3}
        \label{fig:method-comp-dp3}
    \end{subfigure}
    \begin{subfigure}[b]{0.61\textwidth}
        \centering
        \includegraphics[width=\textwidth]{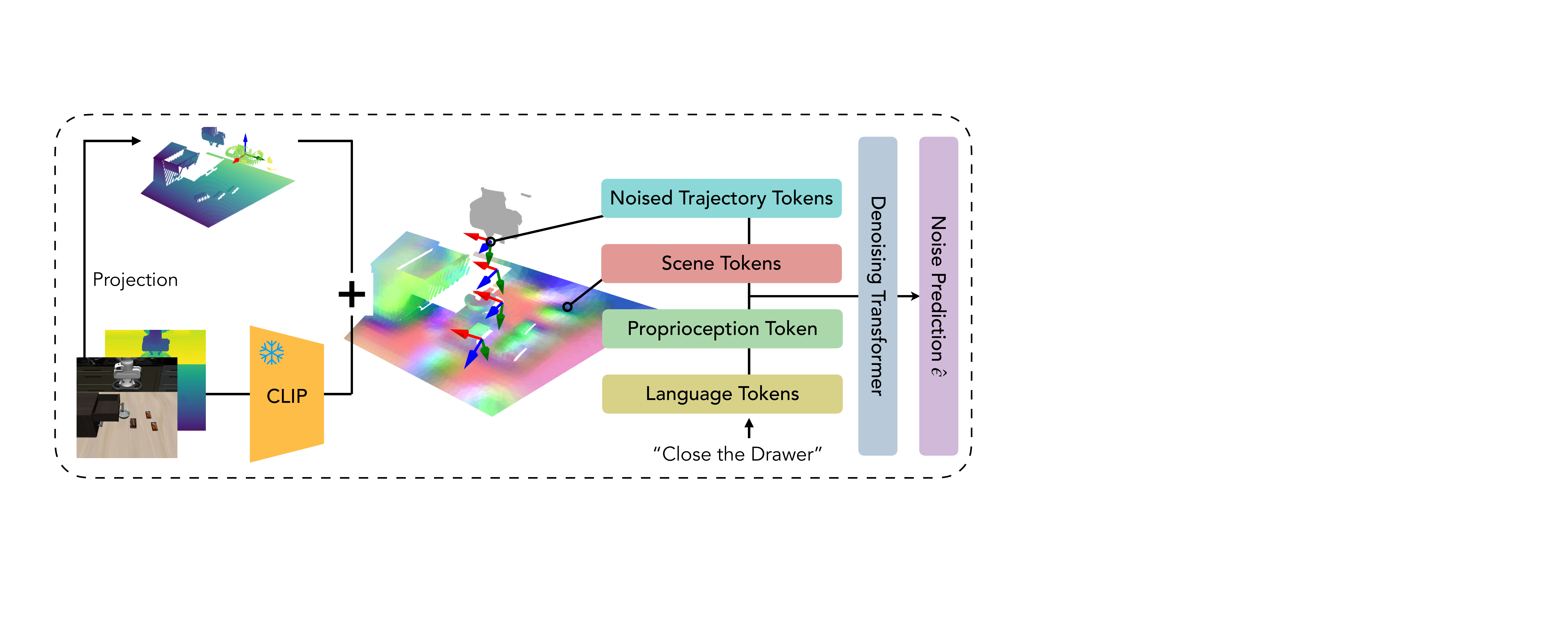}
        \caption{3D Diffuser Actor}
        \label{fig:method-comp-3dda}
    \end{subfigure}

    \begin{subfigure}[b]{0.65\textwidth}
        \centering
        \includegraphics[width=\textwidth]{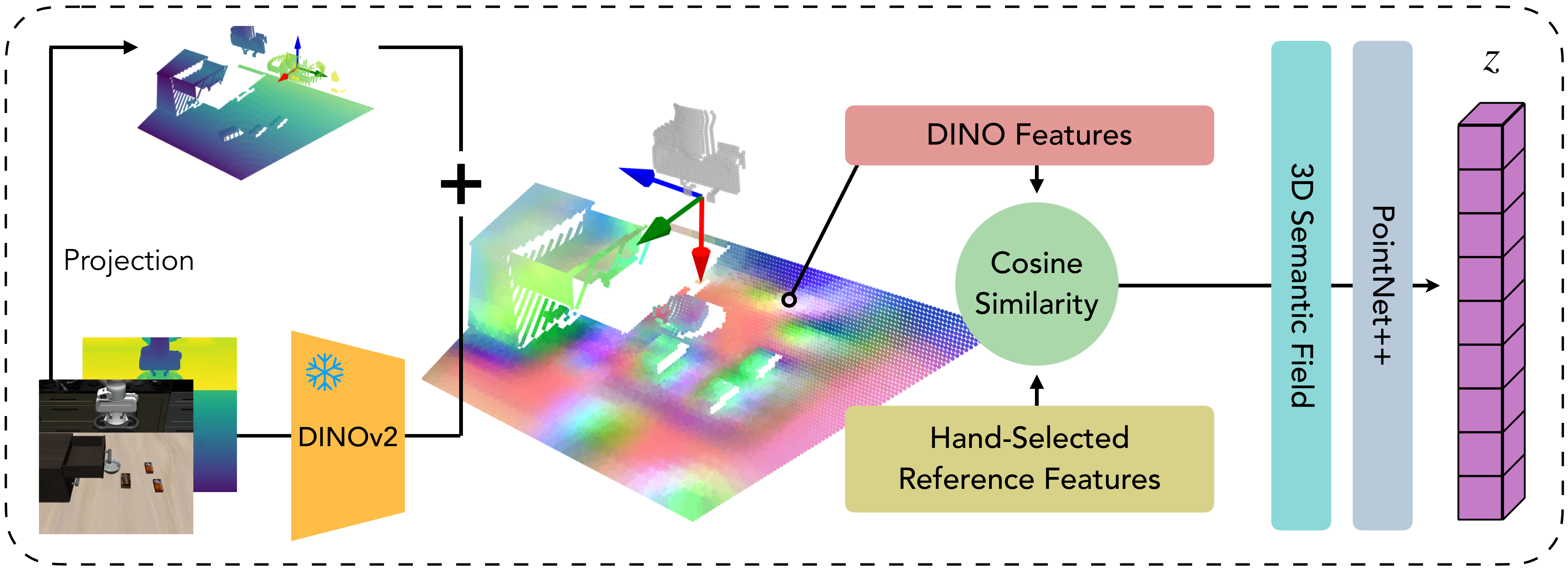}
        \caption{GenDP}
        \label{fig:method-comp-gendp}
    \end{subfigure}

    \caption{In this figure we compare \algabbr{} to several recent methods which also use point clouds for imitation learning. (a) We provide a diagram of \algabbr{} for reference. (b) DP3 \cite{ze2024dp3} omits any semantic information, instead conditioning on colorless point clouds. (c) 3D Diffuser-Actor \cite{3d_diffuser_actor} cross attends between noisy action trajectories and scene tokens. (d) GenDP \cite{wang2023gendp} hand selects important reference features in the training data, and constructs semantic fields through cosine similarity between reference features and scene features.}
    \label{fig:method-comp}
\end{figure*}

In Figure~\ref{fig:method-comp} we present a comparison to several recent imitation learning algorithms which also use point clouds as a means to achieve some degree of generalization.

\textbf{3D Diffusion Policy (DP3)} \cite{ze2024dp3, ze2024idp3} operates based on the core idea that a colorless point cloud contains sufficient information to complete a variety of tasks, and shows impressive results generalizing to new object instances, camera poses and scenes. It lifts the depth maps into colorless point clouds followed by cropping, some amount of downsampling and a neural network to process the points before a max pooling operation.

As we show in our experiments in Sections~\ref{sec:experiments}, these methods fail in multitask and high-precision settings, where reasoning about semantic information regarding the items in the scene is much more important. In multitask settings the semantic information helps the agent to identify its target object, and in high-precision settings the sparse point cloud is insufficient since it does not contain the granularity to discern whether the objects are properly aligned to facilitate the insertion. \algabbr{} differs from these methods not only in its use of semantic information from the CLIP backbone, but also in its point cloud processing steps such as the positional encoding $\gamma$ and the use of the feature space rather than Cartesian coordinates as the metric space for farthest point sampling.

Furthermore, as we show in Section~\ref{sec:real-exps}, these methods are particularly ill-fit to settings with noisy point cloud inputs, such as those arising from stereo depth matching. While \citet{ze2024dp3, ze2024idp3} use a LiDAR camera for their experiments, we argue that the assumption of access to high-quality point clouds is limiting, especially since some large-scale data collection efforts only provide stereo depth (eg DROID \cite{khazatsky2024droid}). By incorporating semantic information in the point cloud, \algabbr{} relies less on accurate point clouds to infer information about the scene and is able to make better sense of these noisy point clouds.

\textbf{3D Diffuser Actor (3DDA)} \cite{3d_diffuser_actor} is another similar method designed to use 3D scene representations to learn a diffusion policy. Specifically, it trains a transformer to, conditioned on the observation and a noisy action $a_t$, predict noise $\epsilon$ which can be removed from $a_t$ to lead it closer to the original action $a_0$. Note that the actions $a_t$ are absolute pose actions.

Like \algabbr{}, it lifts semantic features from a frozen CLIP model into a point cloud. It also lifts the noisy action poses into the cloud, and embeds the proprioceptive information and language instruction into more tokens. Finally, it self attends between points in the scene and cross attends between scene points, action position samples, language tokens and proprioception. 

A core part of this algorithm is that it extracts information from the scene through a self-attention operation on the points in the scene and the noisy trajectory candidate. This has several disadvantages. For one this operation is slow, with time complexity growing quadratically with the number of points in the scene. Furthermore, since this step requires an up-to-date diffusion trajectory candidate, it must be performed again for each diffusion iteration, drastically slowing inference. Next, we believe it is a likely cause of 3DDA's failure to generalize well to unseen camera poses. Because this operation conditions on the 3D geometry of the scene when it self-attends between the points in the scene, it is susceptible to overfitting to that geometry, leading to issues when generalizing to new camera viewpoints.

\textbf{GenDP} \cite{wang2023gendp} is a recent method which extends D$^3$Fields \cite{wang2024d3fields} to general manipulation policies. The key idea is to extract DINOv2 \cite{oquab2023dinov2} features from the images, which are lifted into 3D. It uses the descriptor field algorithm from \citet{wang2024d3fields} to combine features from several viewpoints. Then, it computes cosine similarity between these features and some hand-selected task-relevant reference features (such as shoelaces on a shoe) to compute a semantic feature map. It concatenates these features with Cartesian coordinates to form final representations which it processes with a PointNet++ \cite{qi2017pointnetdeephierarchicalfeature} into a final conditioning vector for a diffusion policy.

Beyond some of the point cloud processing details discussed in Sections~\ref{sec:3d-rep} and \ref{sec:attention}, the key difference between GenDP and \algabbr{} is GenDP's dependence on hand-selected reference features in their construction of the semantic feature map. While this may be acceptable in their setting, where they set out to achieve object instance generalization for specific tasks, this is not scalable to a multitask setting, where it would be necessary to define such reference features for each tasks. \algabbr{} directly uses the outputs of its RGB backbone without need for task-specific reference features.

\clearpage
\section{Extended Method Details}
\label{app:method-detail}

Here we present detailed information about the methods behind the experiments, including detailed information about the inputs to the policies, hyperparameter choices, and action decoder implementation details.

For all algorithms, $\ell_i \in \mathbb{R}^{d_e}$ is the projected output from a pre-trained frozen language encoder (CLIP), $u_{i,t} \in \mathbb{R}^{d_e}$ is the output of a learned proprioception encoding network $U_\theta$ and $z_{i,t} \sim e_\theta(z_{i,t} | o_{i,t})$ is the output of the \algabbr{} encoder. We assume the actions $a_{i,t}$ are captured from an expert demonstrator that consistently completes the task and define $A_{i,t}=\{a_{i, t}, \ldots, a_{i, t+H-1}\}$ to be the length $H$ chunk of actions starting from timestep $i$.

All policies have access to two cameras, as described in Appendix~\ref{sec:exp-details}. All policies have access to proprioceptive information including end effector position and gripper state. All the baseline methods concatenate it and pass it through an MLP encoder, while \algabbr{} uses it to transform the point cloud into the end effector's coordinate frame.

We use the following hyperparameters for all experiments:
\begin{table}[!h]
    \centering
    \begin{tabular}{cc}
        Parameter & Value \\
        \toprule
        Optimizer & Adam \cite{kingma2017adam} \\
        Learning Rate & $1 \times 10^{-4}$ \\
        Weight Decay & $1 \times 10^{-4}$ \\
        Scheduler & Cosine Annealing \cite{loshchilov2017sgdrstochasticgradientdescent} \\
        Batch Size & 64 \\
        Grad Clip & 100 \\
        \# Train Epochs & 100 \\
        Embed Dimension $d_e$ & 256 \\
        \# Downsample Points $p$ & 512 \\
        Weight Initialization & Orthogonal \cite{hu2020provable} \\
        
    \end{tabular}
    \caption{Shared hyperparameters across all experiments}
    \label{tab:shared-hyperparameters}
\end{table}

We found it necessary to change some hyperparameters to accommodate the high precision requirements of the MimicGen environments, and the parameters we tune are presented in Table~\ref{tab:changing-hyperparameters}. In particular, we found finetuning the CLIP backbone to be helpful in reasoning about small position changes needed for the MimicGen experiments. Also we found temporal aggregation, which smoothes predicted trajectories by averaging over several past predictions for the next timestep, to be detrimental in this setting where highly precise motions are required.

\begin{table}[!h]
    \centering
    \begin{tabular}{cccccc}
        Parameter & LIBERO-90 & MG-Coffee & MG-Square & MG-Threading & Real \\
        \toprule
        Temporal Aggregation & True & False & False & False & True \\
        Resample Frequency & N/A & 6 & 2 & 2 & N/A \\
        Finetune CLIP & False & False & True & True & False \\
        
    \end{tabular}
    \caption{Changing hyperparameters across experiments}
    \label{tab:changing-hyperparameters}
\end{table}

\textbf{Action chunking transformer (ACT)} \cite{act} learns to predict chunks of actions by casting the learning as a conditional variational inference problem, and sampling from the likelihood distribution which is parameterized by a transformer model \cite{vaswani2017attention}. Specifically, it learns an encoder $q_\phi(\eta | A_{i,t}, z_{i,t}, u_{i, t})$, where $\eta$ is a latent variable, and decoder $\pi_\phi\left(\hat{A}_{i, t} | z_{i,t}, u_{i, t}, \ell, \eta\right)$ and optimizes the following variational objective:
\begin{multline}
    \mathscr{L}(\phi) = \mathrm{MSE}\left(A_{i, t}, \pi_\phi(z_{i, t}, u_{i,t}, \ell, q_\phi(A_{i,t}, z_{i,t}, u_{i, t}))\right) + \beta D_{\textrm{KL}}(q_\phi(\eta | z_{i,t}, u_{i, t}) || \mathcal{N}(0, 1)).
\end{multline}

In \citet{act}'s implementation, the RGB input is a sequence of patch embeddings. For our implementation we replace this with the output of either the \algabbr{} perception backbone or the backbones described in Section~\ref{sec:baselines}, which we found to simplify the pipeline and still achieve very strong performance. Additionally, for \citet{act}'s implementation the VAE encoder only conditions on proprioceptive information, since it would be too inefficient to also condition on all of the perception outputs. Since our version has far fewer tokens to condition on, the encoder conditions on them without efficiency issues.

For the KL weight we found $\beta=10$ to be good. For baselines we found that a chunk size of $H=15$ worked well and for \algabbr{} we got best results with $H=10$.

\textbf{Diffusion policy (DP)} \cite{diffusionpolicy} samples chunks of actions through a learned reverse Langevin dynamics process from \cite{ho2020denoising}. Specifically, we sample $k \sim \mathrm{Uniform}(\{1, 2, \ldots, K\}), \epsilon \sim \mathcal{N}(0, I)$, and minimize 
\begin{equation} \label{eq:diffusion-denoise-loss}
    \mathscr{L}(\phi) = \mathrm{MSE}(\epsilon, \epsilon_\phi(A_{i,t} + \sigma_t\epsilon, t, z_{i,t}, u_{i,t}, \ell_i))
\end{equation}
where $\sigma_k$ is a noise scale according to the noise schedule and $t$ is the timestep in the noising MDP.

There are several design choices to make in designing a diffusion policy. We used the CNN UNet version of the model with delta pose actions. For conditioning, we concatenate perception, proprioception and language embeddings, which we project down to $d_e$ and use as global conditioning.

We use the DDIM \cite{song2021denoising} scheduler with $100$ training steps and $10$ inference steps and a \texttt{squaredcos\_cap\_v2} beta schedule. For \algabbr{} we found a chunk size of $H=8$ to work well, while a chunk size of $H=16$ worked well for baselines.

\textbf{BAKU} \cite{baku} passes a sequence containing task conditioning $\ell$, observation encodings $z_{i,t}$, proprioception encodings $u_{i,t}$ and a learned readout token to a transformer decoder network $T_\phi$ \cite{vaswani2017attention}. It uses the last output from the transformer as input to an MLP action head $\pi_\phi$, which outputs a Gaussian distribution over chunks of actions. The final pipeline is trained with the following NLL objective:
\begin{equation}
    \mathscr{L}(\phi) = -\log \pi_\phi(A_{i,t} | T_\phi(\ell, z_{i,t}, u_{i,t})[-1]).
\end{equation}

With the exception of chunk size, we use the default hyperparameters from \citet{baku}. For chunk size, we use a value of $H=10$ and $H=15$ for the baselines and \algabbr{} respectively.

\clearpage
\section{Experiment Details}
\label{sec:exp-details}
In this section we provide further details and visualizations describing the experiments described in Section ~\ref{sec:experiments}

\begin{itemize}
    \item \textbf{LIBERO} \cite{liu2024libero} is a multitask learning benchmark consisting of several task suites designed to study lifelong learning. We evaluate on LIBERO-90, which includes 90 rigid-body and articulated manipulation tasks with corresponding natural language instructions. All algorithms have access to a scene camera, wrist camera, and proprioceptive information, and there are 50 demonstrations per task, for a total of 4500 demonstrations including about 670k state-action pairs.
    \item \textbf{MimicGen} \cite{mandlekar2023mimicgen} is a data generation framework that creates large-scale imitation learning datasets from a small number of human demonstrations by segmenting them into object-centric subtasks and generating new trajectories by transforming and executing these segments in novel scenes. It offers a suite of 18 manipulations tasks spanning high-precision, long-horizon, and contact-rich settings. We focus our evaluations on three tasks: \textbf{Coffee}, \textbf{Square}, and \textbf{Threading}.

\end{itemize}

\begin{figure}
    \centering
\includegraphics[width=0.8\textwidth]{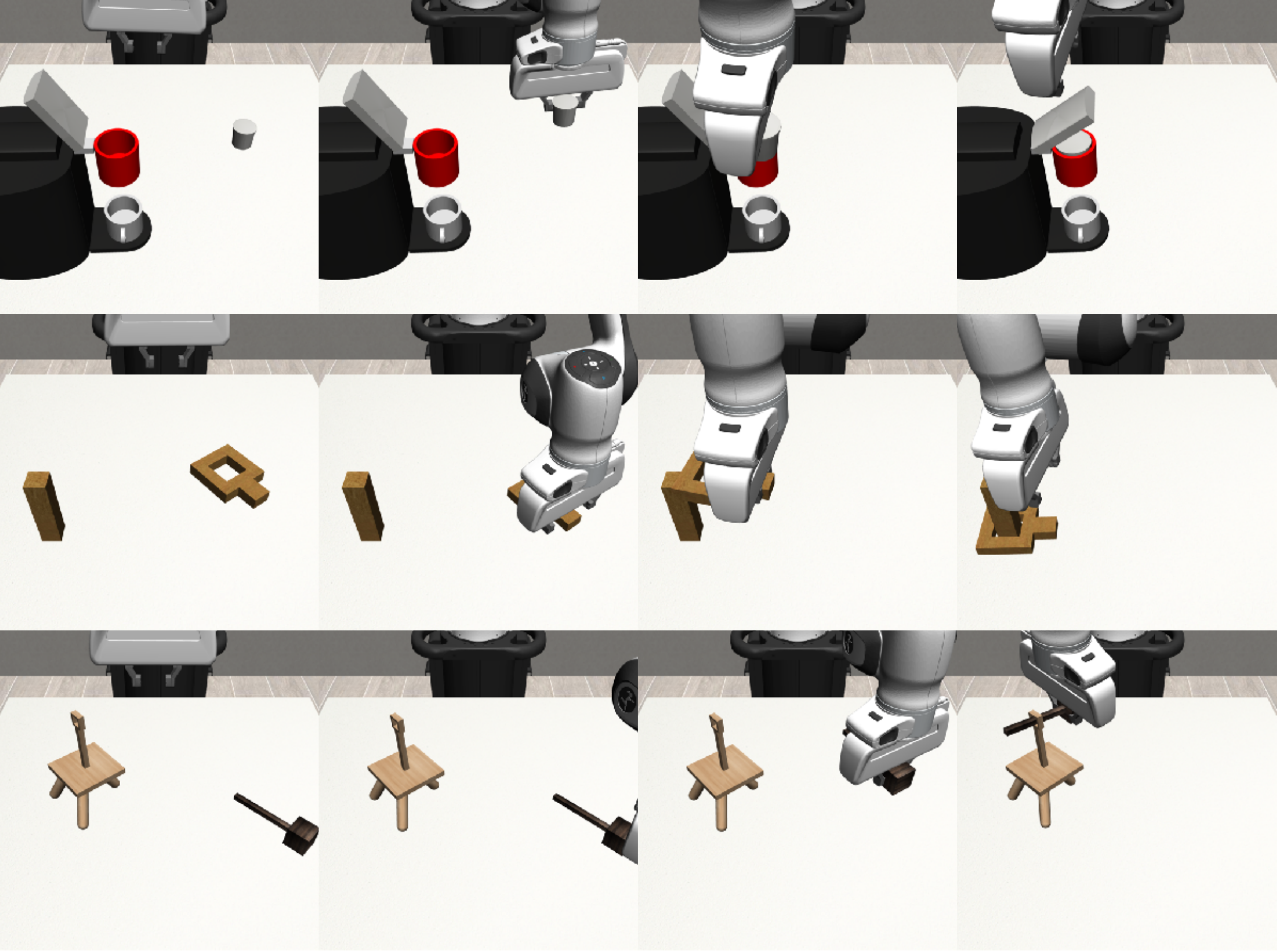}
    \caption{We experiment with the MimicGen Coffee (top), Square (middle) and Threading (bottom) environments, which all require high-precision insertions. For each task we use the D1 reset distribution which requires the policy to learn to complete the task from a wide range of starting configurations.}
    \label{fig:mimicgen_tasks}
\end{figure}

\subsection{LIBERO Environment Details}
\label{app:libero-details}

Most of our experiments use the LIBERO-90 benchmark \cite{liu2024libero}. The benchmark consists of 90 tasks and corresponding language instructions, and is designed to evaluate agents' lifelong learning capabilities. The tasks are spread across 20 scenes in 3 settings (Kitchen, Living Room, Study). 

For all experiments, the action space is delta poses, and the robots are controlled by the Robosuite OSC controller \cite{robosuite2020}.

For the \textbf{change of robot experiment}, we replaced the Franka Panda with the robots shown in Figure~\ref{fig:domain-changes}. Since the robots are controlled by delta pose actions, they share an action space and we could run the outputs from the policy with no further processing. Since we use the gripper joint states as proprioception input, we used the Franka Panda end effector for all robots. In addition to the robots in Figure~\ref{fig:domain-changes} we tried the Sawyer robot, but found it to have low success rates ($<15\%$) for all algorithms.

The LIBERO benchmark comes with 50 initialization states for each task, which specify the poses of the robots and all objects in the scene for deterministic evaluation. For the change of robot experiment, we process these such that all objects in the scene have the original pose, and the end effector of the new robot has the same pose as the original end effector. We found that without this change (ie using the end effector's default initialization pose) all policies had very low success rates, presenting a compelling challenge for future work.

For the \textbf{change of camera experiments} we simply moved the camera and updated the camera calibration matrices. For the results in the main paper we rotate the camera about the vertical axis going through the end-effector starting position, as described in Section~\ref{sec:unseen-cam}. For the tables in the appendix, we consider \texttt{small}, \texttt{medium} and \texttt{large} changes in camera pose. 
Specifically, for the \texttt{small} change, we shift the camera by $(0.0, +0.3, -0.1)\, \text{m}$, for the \texttt{medium} change, we shift the camera by $(-0.2, +0.7, -0.2)\, \text{m}$ and for the \texttt{large} change, we shift the camera by $(-1.2, +1.0, -0.2)\, \text{m}$. For all camera position changes we subsequently rotate the camera such that it is facing the end effector's starting location.

\subsection{MimicGen Environment Details}

For the MimicGen experiments we use the provided D1 datasets, each of which includes 1000 trajectories procedurally generated from 10 human trajectories. We choose the Coffee, Square and Threading tasks because they are difficult enough to showcase \algabbr{}'s applicability to difficult high-precision insertion tasks.

We use the same change of robot and change of camera pose experiments from LIBERO. We generate initial states for each environment that we use to ensure fair evaluation.

\subsection{Real Experiment Details}
\label{app:real-details}

The real benchmark includes 6 pick-and-place tasks with between 59 and 87 demonstrations per task for an average of 73 demonstrations per task, collected using a Meta Quest 3 headset for teleoperation. A sample of tasks are shown in Figure~\ref{fig:teaser} and the full suite is shown in the supplement. We use a Universal Robots UR5 and a Robotiq 2F-85 gripper. The setup includes two Realsense D435 scene cameras and a Realsense D405 wrist camera. A diagram of the setup is shown in Figure~\ref{fig:real-diagram}.

For all observation backbones we train diffusion policies with absolute end-effector pose action space. All policies are trained in a multitask fashion using CLIP-encoded language instructions to differentiate tasks. 

\begin{figure}
    \centering
    \includegraphics[width=\textwidth]{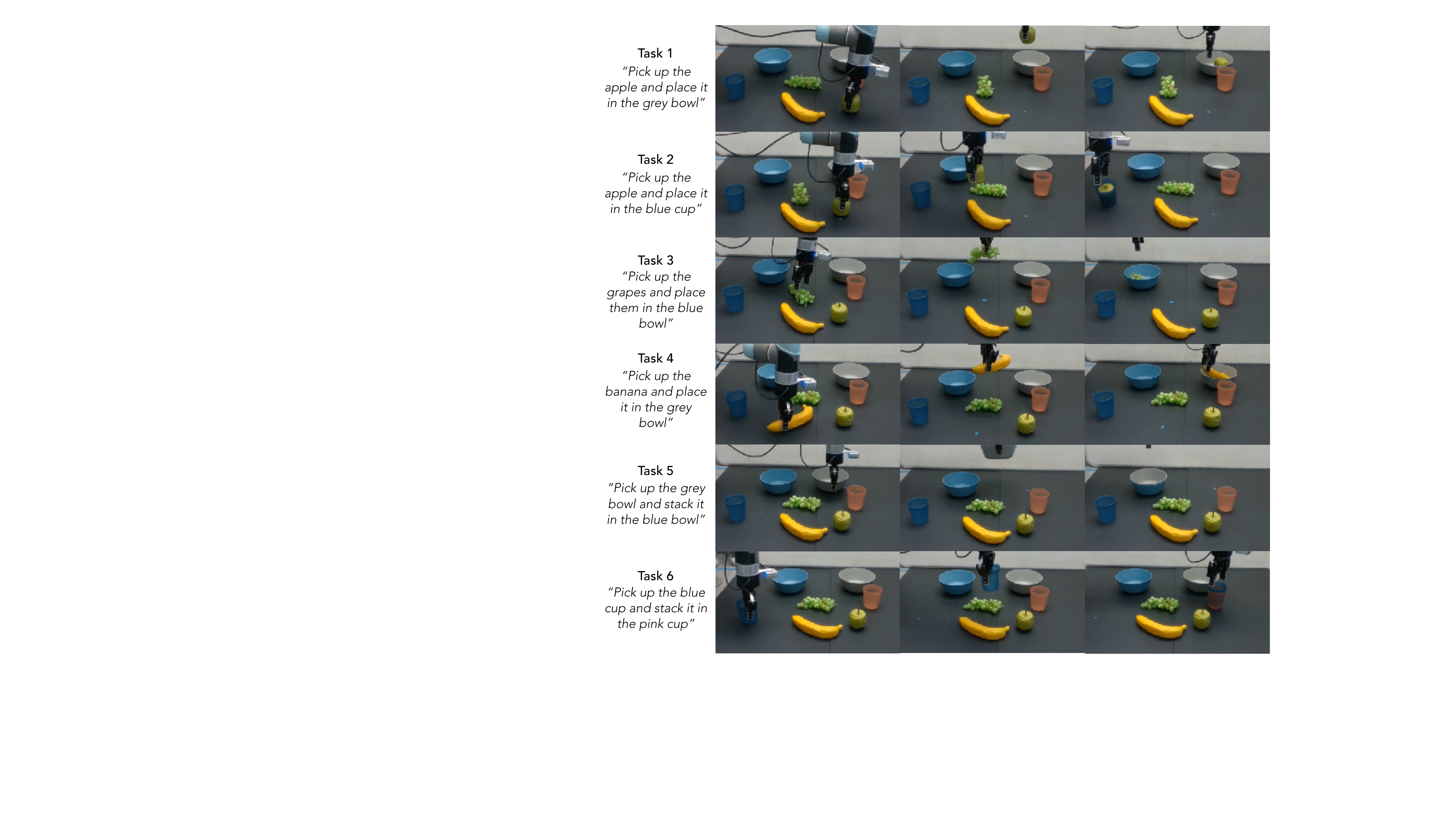}
    \caption{Our real-world multitask imitation learning benchmark includes 6 pick-and-place tasks shown above.}
    \label{fig:real_tasks}
\end{figure}

For our results we record \textit{progress towards task completion} as opposed to raw success rates, where grasping the target object corresponds to $\frac{1}{3}$ progress, bringing it into some contact with the goal object corresponds to $\frac{2}{3}$ and completing the task is $\frac{3}{3}$. We train two random seeds per policy and perform 5 evaluation trials per seed for a total of 10 trials per task per algorithm.

Our dataset includes a total of 437 trajectories for 6 tasks. In Table \ref{tab:real-dataset} we present exact statistics about the datasets for the 6 tasks. In Figure~\ref{fig:real_tasks} we show visualizations of all tasks present in our multitask imitation learning benchmark.

For all policies we use end effector pose as proprioceptive input. For all objects we select a home location and initialize them within a 10 cm radius of that home location. For objects which are not axisymmetric about the vertical axis (eg. the banana and grapes) we randomly rotate them. At evaluation time we deterministically initialize object poses in a grid near the home location.

\begin{table}[!t]
    \centering
    \begin{tabular}{c|c|cc}
        Task No. & Language Instruction & \# Trajectories & \# Samples \\
        \midrule
        1 & ``Pick up the apple and place it in the grey bowl'' & 87 & 19209 \\
        2 & ``Pick up the apple and place it in the blue cup'' & 80 & 15925 \\
        3 & ``Pick up the grapes and place them in the blue bowl'' & 59 & 10339 \\
        4 & ``Pick up the banana and place it in the grey bowl'' & 74 & 14635 \\
        5 & ``Pick up the grey bowl and stack it in the blue bowl'' & 60 & 10775 \\
        6 & ``Pick up the blue cup and stack it in the pink cup'' & 77 & 16583 \\
        \midrule
        \multicolumn{2}{r}{Total} & 437 & 87466 \\
    \end{tabular}
    \caption{Statistics about our real dataset.}
    \label{tab:real-dataset}
\end{table}

\clearpage
\section{Further Results and Visualizations}
\label{app:sim-results}

In this section we present additional results that did not fit in the main paper. For all experiments we consider the LIBERO-90 benchmark.

\subsection{Full LIBERO Results}

Here we present tables with full numerical results that were summarized in the bar charts in the paper.

Table ~\ref{table:original-results} shows the in-distribution results for all three action decoders on the LIBERO-90 dataset. We see that \algabbr{} achieves strong multitask performance in this setting and combines effectively with each action decoder.

\begin{table}[h!]
\centering
\caption{Success rates for in-distribution evaluations on the LIBERO 90 benchmark. We see that \algabbr{} has a similar modeling capacity to strong baselines such as RGB and 3DDA for LIBERO 90.}
\label{table:original-results}
\begin{tabular}{lccc}
\toprule
Encoder & ACT & Diffusion Policy & BAKU \\
\midrule
RGB & $\mathbf{0.908}$ & $\mathbf{0.905}$ & $0.914$ \\
RGBD & $0.705$ & $0.868$ & $0.789$ \\
DP3 & $0.753$ & $0.687$ & $0.681$ \\
3D Diffuser Actor & - & 0.837 & - \\
\algabbr & $0.903$ & $0.880$ & $\mathbf{0.919}$ \\
\bottomrule
\end{tabular}
\end{table}

Table~\ref{table:robot-results} presents full results for the cross embodiment experiment on LIBERO-90. We see that \algabbr{} combines will with ACT and BAKU to achieve strong cross embodiment results across the borad for these algorithms. When combined with DP, it achieves strong performance when switching to the UR5e embodiment, but is outperformed by 3DDA, which also uses a 3D scene representation that is more robust to changes in embodiment.

\begin{table}[h!]
\centering
\caption{Success rates when rolling out policies in a zero-shot cross-embodiment setting. We see that \algabbr{} consistently outperforms all other observation encoders and achieves similar performance to 3DDA.}
\label{table:robot-results}
\begin{tabular}
{lccc}
\toprule
Algorithm & UR5e & Kinova3 & IIWA \\
\midrule
ACT + RGB & $0.578$ & $0.643$ & $0.525$ \\
ACT + RGBD & $0.493$ & $0.505$ & $0.420$ \\
ACT + DP3 & $0.592$ & $0.574$ & $0.613$ \\
ACT + \algabbr & $\mathbf{0.777}$ & $\mathbf{0.743}$ & $\mathbf{0.714}$ \\
\midrule
DP + RGB & $0.543$ & $0.562$ & $0.415$ \\
DP + RGBD & $0.464$ & $0.441$ & $0.332$ \\
DP + DP3 & $0.563$ & $0.404$ & $0.423$ \\
3D Diffuser Actor & $0.722$ & $\mathbf{0.759}$ & $\mathbf{0.651}$ \\
DP + \algabbr & $\mathbf{0.757}$ & $0.529$ & $0.505$ \\
\midrule
BAKU + RGB & $0.442$ & $0.458$ & $0.352$ \\
BAKU + RGBD & $0.419$ & $0.356$ & $0.310$ \\
BAKU + DP3 & $0.579$ & $0.468$ & $0.511$ \\
BAKU + \algabbr & $\mathbf{0.801}$ & $\mathbf{0.653}$ & $\mathbf{0.718}$ \\
\bottomrule
\end{tabular}
\end{table}

In Table~\ref{table:camera-change} we present results for all algorithms under a zero-shot camera change experiment, with camera viewpoints described in Section~\ref{app:libero-details}. Note that these angles are different than those presented in the main body of the paper. We see that, across all action decoders, \algabbr{} enables strong performance with unseen camera poses.

\begin{table}[!h]
\centering
\caption{Success rates when rolling out policies in a zero-shot camera change setting for LIBERO-90. We see that \algabbr{} consistently outperforms all comparisons and achieves success rates an average of $56.1\%$ higher than the next best comparison.}
\setlength{\tabcolsep}{3pt} 
\label{table:camera-change}
\begin{tabular}{lccc}
\toprule
Algorithm & Small & Medium & Large \\
\midrule
ACT + RGB & $0.310$ & $0.216$ & $0.230$ \\
ACT + RGBD & $0.323$ & $0.220$ & $0.175$ \\
ACT + DP3 & $0.376$ & $0.218$ & $0.081$ \\
ACT + \algabbr & $\mathbf{0.787}$ & $\mathbf{0.706}$ & $\mathbf{0.749}$ \\
\midrule
DP + RGB & $0.133$ & $0.058$ & $0.042$ \\
DP + RGBD & $0.178$ & $0.075$ & $0.080$ \\
DP + DP3 & $0.179$ & $0.174$ & $0.028$ \\
3D Diffuser Actor & $0.226$ & $0.208$ & $0.186$ \\
DP + \algabbr & $\mathbf{0.740}$ & $\mathbf{0.798}$ & $\mathbf{0.791}$ \\
\midrule
BAKU + RGB & $0.141$ & $0.061$ & $0.043$ \\
BAKU + RGBD & $0.176$ & $0.065$ & $0.068$ \\
BAKU + DP3 & $0.311$ & $0.261$ & $0.135$ \\
BAKU + \algabbr & $\mathbf{0.847}$ & $\mathbf{0.829}$ & $\mathbf{0.829}$ \\
\bottomrule
\end{tabular}
\end{table}

\subsection{Real}

In Table~\ref{table:real-mt} and Table~\ref{table:real-zero-shot} we present results for all tasks in our real-world multitask benchmark. We see that in the original domain, \algabbr{} achieves the strongest overall performance in two of the six tasks. When transfering to the unseen camera pose, \algabbr{} achieves the strongest performance by far and is the only algorithm to consistently complete the tasks.

\begin{table}[!h]
\centering
\caption{Average task completion for each task in our benchmark when using the original camera viewpoint.}
\setlength{\tabcolsep}{3pt} 
\label{table:real-mt}

\begin{tabular}
{l|cccccc|r}
\toprule
Algo. & Task 1 & Task 2 & Task 3 & Task 4 & Task 5 & Task 6 & Avg.\\
\midrule
RGB & $\mathbf{100}$ & $90.0$ & $43.3$ & $50.0$ & $50.0$ & $\mathbf{100}$ & $72.2$ \\ 
DP3 & $0.0$ & $0.0$ & $53.3$ & $3.3$ & $26.7$ & $0.0$ & $13.9$ \\ 
3DDA & $\mathbf{100}$ & $\mathbf{100}$ & $90.0$ & $\mathbf{70.0}$ & $\mathbf{86.7}$ & $\mathbf{100}$ & $\mathbf{91.1}$ \\ 
\algabbr{} & $\mathbf{100}$ & $90.0$ & $\mathbf{93.3}$ & $63.3$ & $80.0$ & $83.3$ & $85.0$ \\

\bottomrule
\end{tabular}
\end{table}

\begin{table}[!h]
\centering
\caption{Average task completion for each task in our benchmark when rolling out with the unseen camera viewpoint shown in Figure~\ref{fig:real-test-view}.}
\setlength{\tabcolsep}{3pt} 
\label{table:real-zero-shot}

\begin{tabular}
{l|cccccc|r}
\toprule
Algo. & Task 1 & Task 2 & Task 3 & Task 4 & Task 5 & Task 6 & Avg.\\
\midrule
 
RGB & $30.0$ & $16.7$ & $50.0$ & $0.0$ & $33.3$ & $36.7$ & $27.8$ \\ 
DP3 & $0.0$ & $10.0$ & $23.3$ & $0.0$ & $16.7$ & $0.0$ & $8.3$ \\ 
3DDA & $80.0$ & $26.7$ & $50.0$ & $20.0$ & $33.3$ & $3.3$ & $35.6$ \\ 
\algabbr{} & $\mathbf{100}$ & $\mathbf{83.3}$ & $\mathbf{90.0}$ & $\mathbf{53.3}$ & $\mathbf{73.3}$ & $\mathbf{76.7}$ & $\mathbf{79.4}$ \\

\bottomrule
\end{tabular}
\end{table}

\subsection{Attention Maps}

\begin{figure}
    \centering
    \includegraphics[trim=150 100 150 100, clip, width=0.24\textwidth]{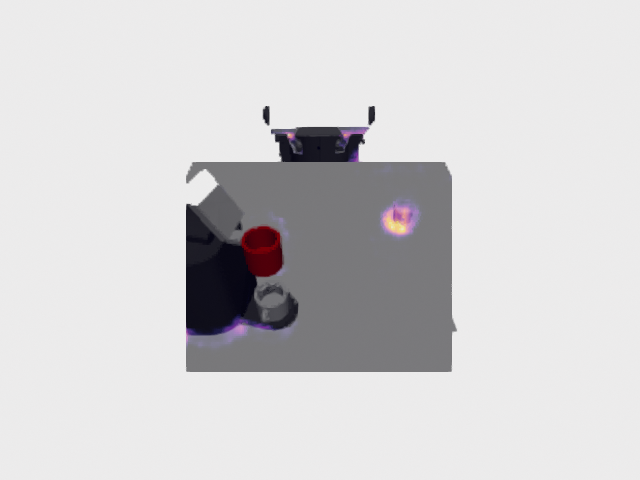}
    \includegraphics[trim=150 100 150 100, clip, width=0.24\textwidth]{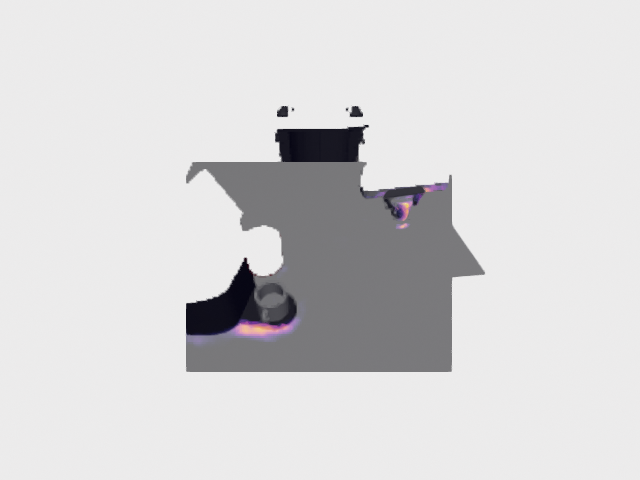}
    \includegraphics[trim=150 100 150 100, clip, width=0.24\textwidth]{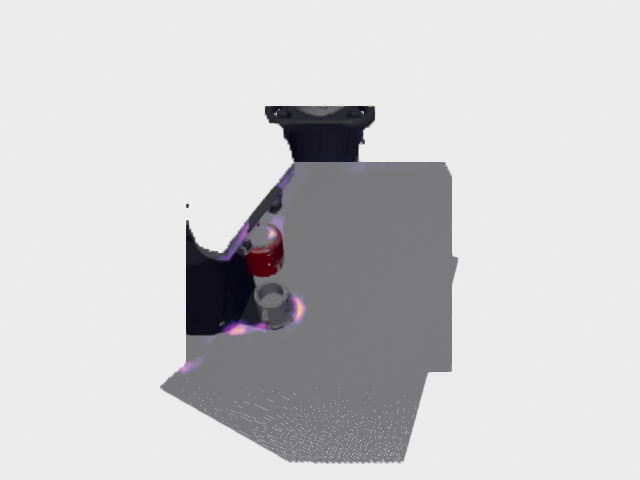}
    \includegraphics[trim=150 100 150 100, clip, width=0.24\textwidth]{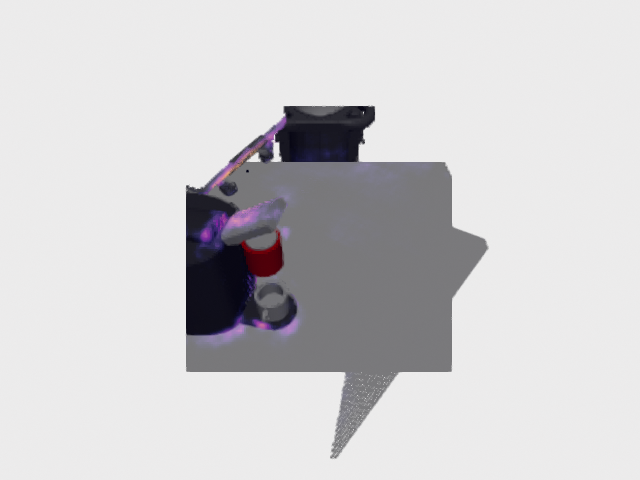}
    \includegraphics[trim=150 100 150 100, clip, width=0.24\textwidth]{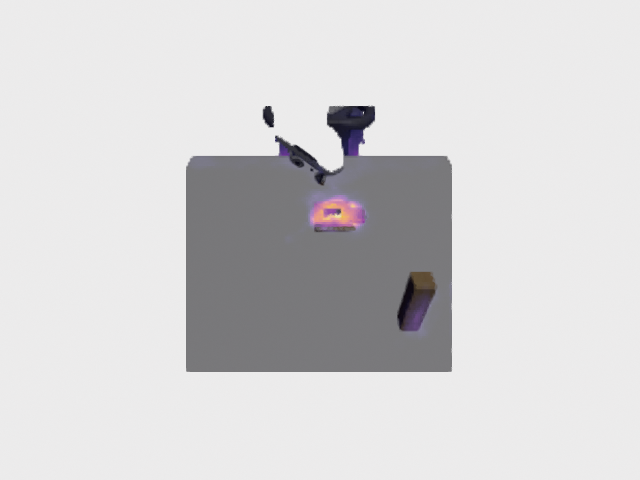}
    \includegraphics[trim=150 100 150 100, clip, width=0.24\textwidth]{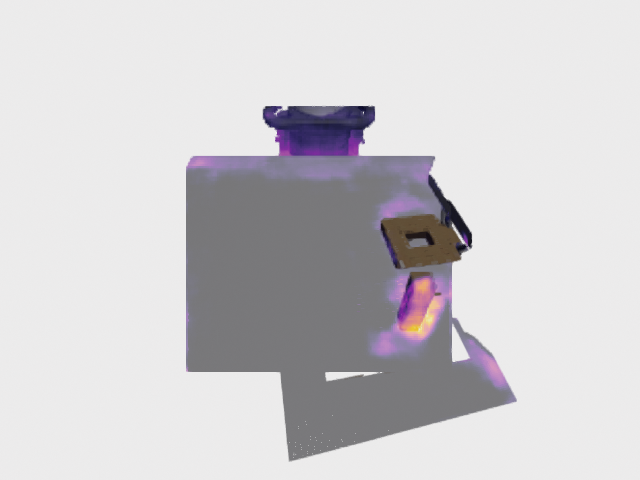}
    \includegraphics[trim=150 100 150 100, clip, width=0.24\textwidth]{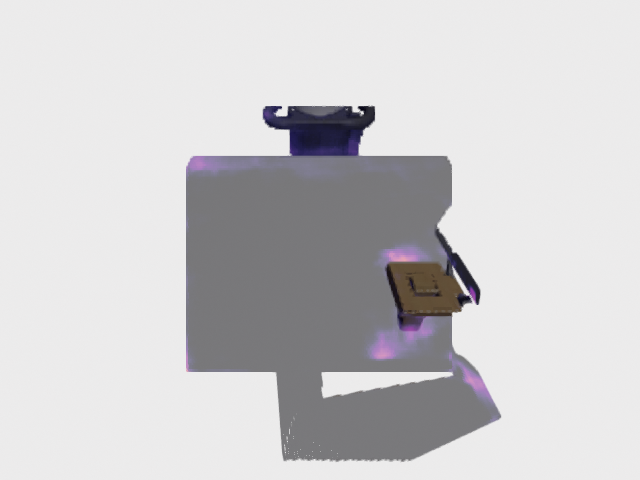}
    \includegraphics[trim=150 100 150 100, clip, width=0.24\textwidth]{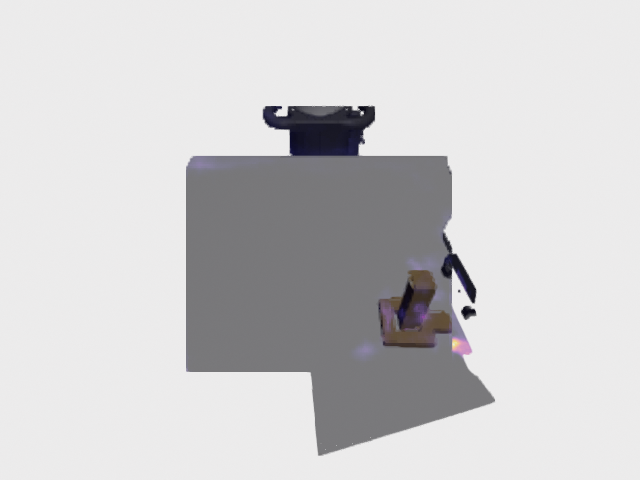}
    \includegraphics[trim=150 100 150 100, clip, width=0.24\textwidth]{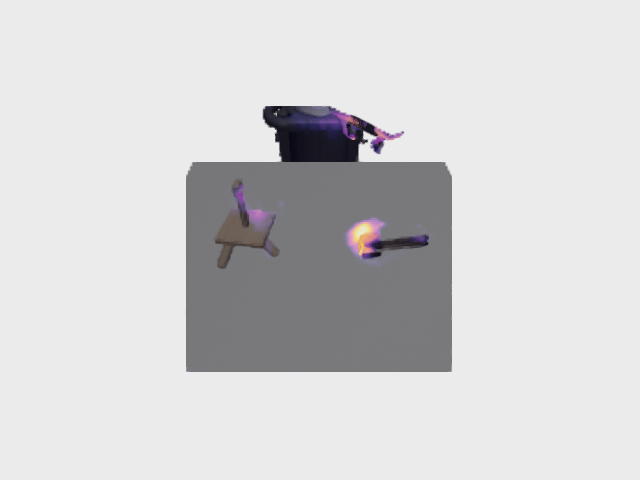}
    \includegraphics[trim=150 100 150 100, clip, width=0.24\textwidth]{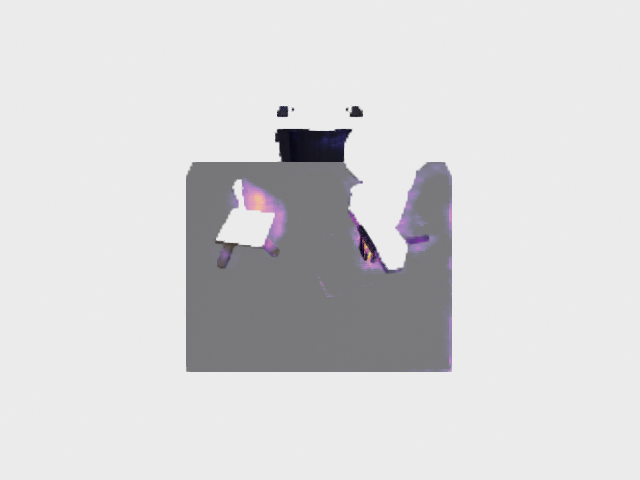}
    \includegraphics[trim=150 100 150 100, clip, width=0.24\textwidth]{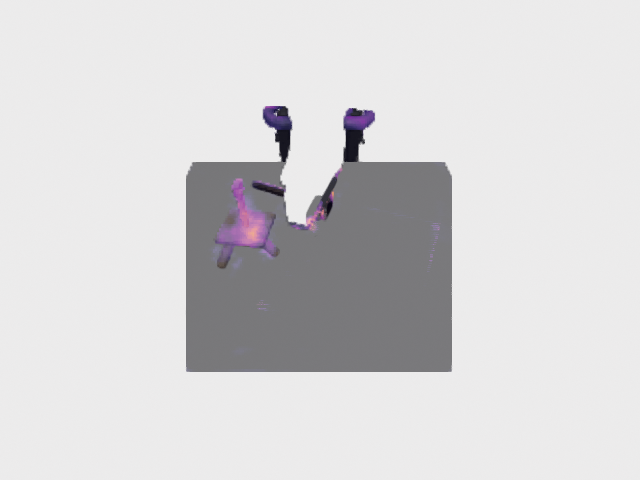}
    \includegraphics[trim=150 100 150 100, clip, width=0.24\textwidth]{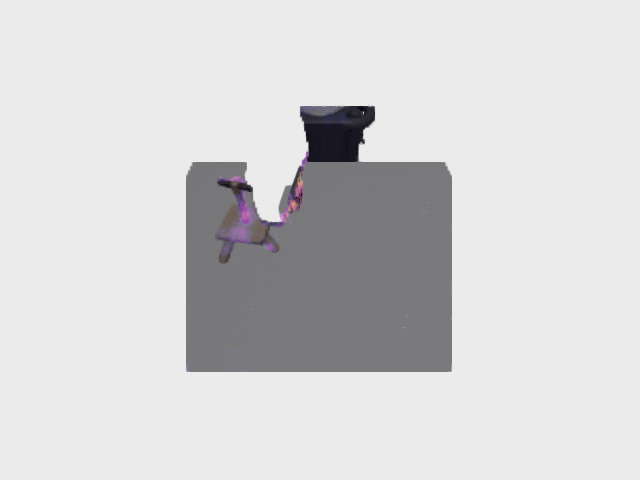}
    \caption{We study attention maps from the MimicGen environments at various critical points in the rollouts. In order to create attention maps at full resolution we bilinearly interpolate feature maps to the size of the full point cloud, which causes them to be slightly fuzzy around the edges of the objects.}
    \label{fig:attn-maps}
\end{figure}

One unique attribute from our architecture is the use of attention pooling, which allows us to visualize attention maps and understand which parts of the scene our policy attends to at which times, as we do in Figure~\ref{fig:attn-maps}.

For the Coffee task (top) we see that early on it attends primarily to the coffee pod, which it is trying to grasp. Later on it attends to points around the base of the coffee machine, which it uses to localize itself with respect to the machine for the precise insertion. For the Square task (middle) it begins by attending primarily to the object it must grasp before later attending to the peg. For the Threading task (bottom), it once again starts by attending to the object it must grasp, and then attends to the tripod to help with aligning for the insertion.

\subsection{Architecture Study}
\label{sec:arch-variation}

In this section we study design choices in the final part of our pipeline, which reduces the point cloud constructed in Section~\ref{sec:3d-rep} into a single vector for use as conditioning for the action decoder. Overall we conclude that this design choice is not as important as the design choices discussed earlier in the paper.

Our pipeline uses an attention pooling reduction as described in Section~\ref{sec:attention}. We compare to the following architectures:
\begin{itemize}
    \item The architecture from \textbf{DP3} \cite{ze2024dp3}, which runs each point through an MLP before performing max pooling across the point cloud. This method is similar to ours, but replacing the attention pooling with max pooling.
    \item The architecture from \textbf{GenDP} \cite{wang2023gendp}, which uses PointNet++ \cite{qi2017pointnetdeephierarchicalfeature}, a hierarchical point processing architecture.
\end{itemize}

\begin{table}[htbp]
\centering
\caption{How do choices in the embedding extractor architecture affect the final success rate? In this table we present success rates after varying the architecture as discussed in Section~\ref{sec:arch-variation}. While \algabbr{} has narrowly better overall performance, the results are inconclusive. All experiments are from combining \algabbr{} with BAKU. 
}
\vspace{-0.05in}
\label{table:arch-ablation}

\begin{tabular}
{lcccc}
\toprule
 & DP3 & iDP3 & GenDP & Ours \\
\midrule
Orig. & $\mathbf{0.920}$ & $0.881$ & $0.886$ & $0.919$ \\
\midrule
UR5e & $\mathbf{0.827}$ & $0.758$ & $0.735$ & $0.801$ \\
Kinova3 & $\mathbf{0.700}$ & $0.611$ & $0.619$ & $0.653$ \\
IIWA & $0.695$ & $0.660$ & $0.697$ & $\mathbf{0.718}$ \\
\midrule
Small & $0.713$ & $0.747$ & $\mathbf{0.859}$ & $0.847$ \\
Medium & $0.641$ & $0.781$ & $0.792$ & $\mathbf{0.829}$ \\
Large & $0.649$ & $0.747$ & $0.781$ & $\mathbf{0.829}$ \\
\bottomrule
\end{tabular}

\end{table}

We present results from this comparison in Table~\ref{table:arch-ablation}. Overall, we see that although \algabbr{} narrowly outperforms the comparison architectures, the results do not resoundingly indicate any one architecture is best. This indicates to us that for future work it may be more fruitful to study changes to the structure of the point cloud itself rather than the final embedding extractor architecture.

\subsubsection{RGB Backbone Choice}

In this section we study the choice of RGB backbone used to define $f$ in Section~\ref{sec:3d-rep}. In the final version of \algabbr{} we use a CLIP \cite{clip} backbone with frozen weights, similarly with \cite{3d_diffuser_actor}. However, it is unclear whether pretrained CLIP weights are indeed the best choice. To that end, in Table~\ref{table:backbones} we compare CLIP weights with pretrained ResNet weights \cite{resnet}, and ablate the finetuning. We also explore the use of DINOv2 \cite{oquab2023dinov2} in Table~\ref{table:dino}, which is known to extract richer local features important for robotic manipulation tasks \cite{wang2023gendp, wang2024d3fields}.

First of all, we see no clear difference between performance with CLIP weights and ResNet18 weights. This is surprising, and indicates to us that the model is likely not reasoning deeply about the semantic information in $F$, but rather is using the feature vectors to look up important points.

Another interesting takeaway from Table~\ref{table:backbones} is that finetuning does not help with in-distribution settings but is harmful to out-of-distribution settings, especially when transferring to novel camera viewpoints. One explanation is that finetuning leads the backbone to overfit to the dataset and fail more often for slightly OOD states in in-distribution settings and fail frequently for OOD evaluation settings.

\begin{table}[!h]
\centering
\caption{How does the choice of image feature extractor backbone affect the performance of \algabbr{}? `FT' means the model was finetuned while `Fr' means its weights are frozen. `RN18' means a ResNet-18 pretrained with ImageNet weights \cite{resnet, deng2009imagenet}, and CLIP refers to CLIP \cite{clip}. All experiments are from the combination of \algabbr{} and BAKU.}
\vspace{-0.05in}
\label{table:backbones}

\begin{tabular}
{lcccc}
\toprule
 & FT RN18 & Fr RN18 & FT CLIP & Fr CLIP \\
\midrule
Orig. & $0.909$ & $0.906$ & $0.912$ & $\mathbf{0.919}$ \\
\midrule
UR5e & $0.774$ & $0.758$ & $0.774$ & $\mathbf{0.801}$ \\
Kinova3 & $0.597$ & $0.615$ & $0.611$ & $\mathbf{0.653}$ \\
IIWA & $0.674$ & $0.704$ & $0.654$ & $\mathbf{0.718}$ \\
\midrule
Small & $0.840$ & $\mathbf{0.848}$ & $0.819$ & $0.847$ \\
Medium & $0.819$ & $\mathbf{0.840}$ & $0.806$ & $0.829$ \\
Large & $0.807$ & $0.822$ & $0.780$ & $\mathbf{0.829}$ \\
\bottomrule
\end{tabular}

\vspace{-0.1in}
\end{table}

Finally, in Table~\ref{table:dino}, we explore the potential of DINOv2 \cite{oquab2023dinov2} as a richer feature extractor. DINOv2 is known for capturing detailed local features, as a result of its self-supervised pretraining task, which emphasizes learning spatially localized representations. This ability could enhance the model's performance at identifying fine-grained details critical for robotic manipulation tasks. For our experiments, we use the \texttt{dinov2\_small} model, which outputs a 384-dimensional feature vector for each patch. We modify the stride in the patch extractors and pad the input images to ensure the feature volume output from DINOv2 is compatible with the \algabbr{} architecture.

We investigate the usefulness of representations from different layers of DINOv2 by extracting features from various depths:

\begin{itemize}
    \item \textbf{Patch Embeddings}: The embeddings from the patch projection before being passed into the DINOv2 backbone.
    \item \textbf{Layers $n$}: The feature vectors output by the DINOv2 backbone after being sequentially processed through the first $n$ transformer blocks.
\end{itemize}

In all cases, we discard the class token (\texttt{[CLS]}) to focus on the local features. We avoid experimenting with deeper layers or finetuning as we found DINOv2 feature extraction to be prohibitively time consuming and processing beyond the third layer becomes impractical for both training time and inference speed.

In the original setting, there is no noticeable difference in performance when using DINOv2 features compared to CLIP or ResNet. This consistency across RGB backbones reinforces our earlier claim that the model primarily uses the features to identify task-relevant points rather than leveraging their semantic richness.

In the \textbf{change of robot experiments}, DINOv2 features show a slight advantage, with a $2\text{-}3\%$ improvement, particularity in the shallowest layer (Layer 1). This may suggest that the local features extracted by DINOv2's shallow layers are less sensitive to the agent's embodiment and more focused on the objects and interactions in the scene, enabling better generalization to new robot embodiments. 

In the \textbf{change of camera experiments} DINOv2 under-performed compared to CLIP or ResNet, likely because the shallow layers in DINOv2 focus on local details and fail to capture the broader spatial context needed for viewpoint changes. CLIP, on the other hand, is better at capturing global information, making it more effective for scenarios involving significant camera shifts. Notably, performance improves when using features from deeper DINOv2 layers, which are known to encode information over larger spatial extents \cite{oquab2023dinov2}. This highlights the importance of global representations for handling viewpoint changes, as they allow the model to reason about the overall structure of the scene rather than relying solely on local features.

\begin{table}[!h]
\centering
\caption{Effect of using different layers in DINOv2 on the performance of \algabbr{}. `Patch' refers to the initial patch embeddings before the DINOv2 backbone, while `L1', `L2', and `L3' denote features extracted after the first, second, and third transformer block, respectively. `Ours' corresponds to the baseline model, \algabbr{} using the CLIP backbone with frozen weights. All experiments are from the combination of \algabbr{} and BAKU. Note that the versions with DINO features are from an older version of our pipeline which didn't use proprioception, which explains their marked deficiency for new camera poses.}
\label{table:dino}

\begin{tabular}{lccccc} 
\toprule
& Patch & L1 & L2 & L3 & Ours \\ 
\midrule 
Orig. & $0.909$ & $\mathbf{0.937}$ & $0.929$ & $0.923$ & $0.919$ \\ 
\midrule 
UR5e & $0.809$ & $\mathbf{0.840}$ & $0.818$ & $0.820$ & $0.801$ \\ 
Kinova3 & $0.714$ & $0.766$ & $\mathbf{0.774}$ & $0.713$ & $0.653$ \\ 
IIWA & $0.662$ & $0.716$ & $0.697$ & $0.711$ & $\mathbf{0.718}$ \\ 
\midrule 
Small & $0.442$ & $0.549$ & $0.543$ & $0.591$ & $\mathbf{0.847}$ \\ 
Medium & $0.208$ & $0.426$ & $0.513$ & $0.585$ & $\mathbf{0.829}$ \\ 
Large & $0.167$ & $0.437$ & $0.558$ & $0.654$ & $\mathbf{0.829}$ \\ 
\bottomrule 
\end{tabular}
\vspace{-0.1in}
\end{table}

\subsubsection{Cropping}

\begin{figure*}
    \centering
    \begin{subfigure}[b]{0.3\textwidth}
        \centering
        \includegraphics[width=0.7\textwidth]{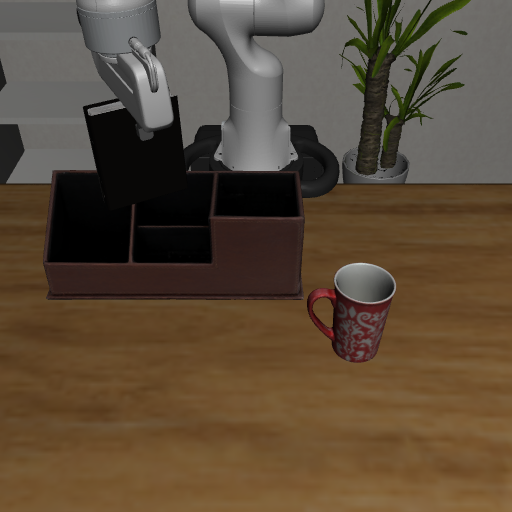}
        \caption{RGB Image}
        \label{fig:crop-rgb}
    \end{subfigure}
    \begin{subfigure}[b]{0.3\textwidth}
        \centering
        \includegraphics[width=\textwidth]{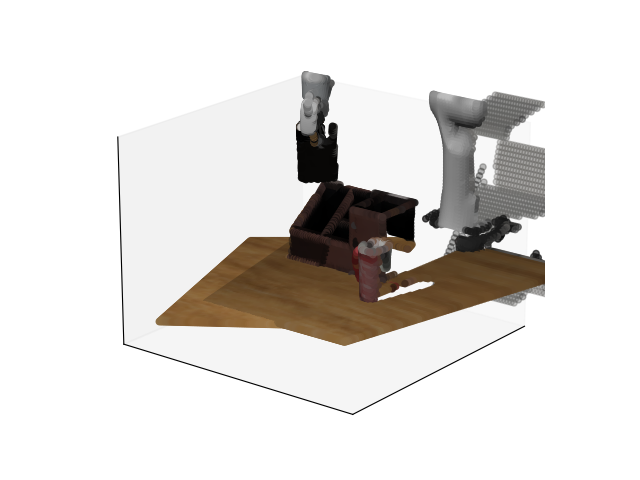}
        \caption{Uncropped}
        \label{fig:crop-uncropped}
    \end{subfigure}
    
    \begin{subfigure}[b]{0.3\textwidth}
        \centering
        \includegraphics[width=\textwidth]{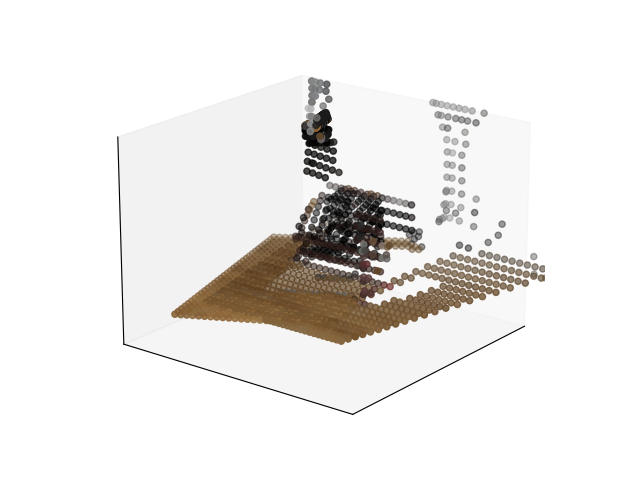}
        \caption{Loose cropping}
        \label{fig:crop-loose}
    \end{subfigure}
    \begin{subfigure}[b]{0.3\textwidth}
        \centering
        \includegraphics[width=\textwidth]{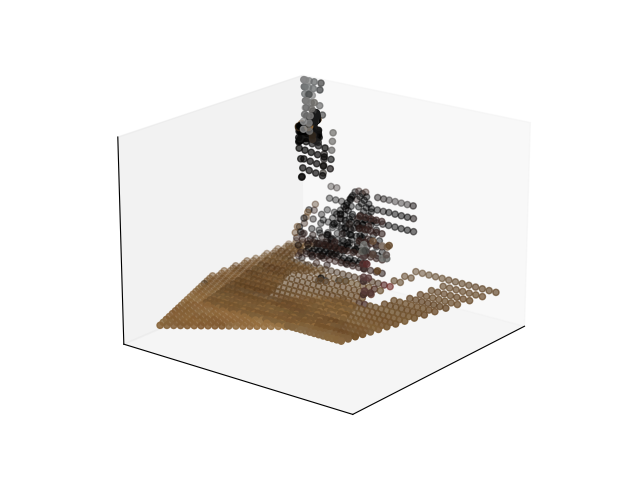}
        \caption{Tight Cropping}
        \label{fig:crop-tight}
    \end{subfigure}
    \begin{subfigure}[b]{0.3\textwidth}
        \centering
        \includegraphics[width=\textwidth]{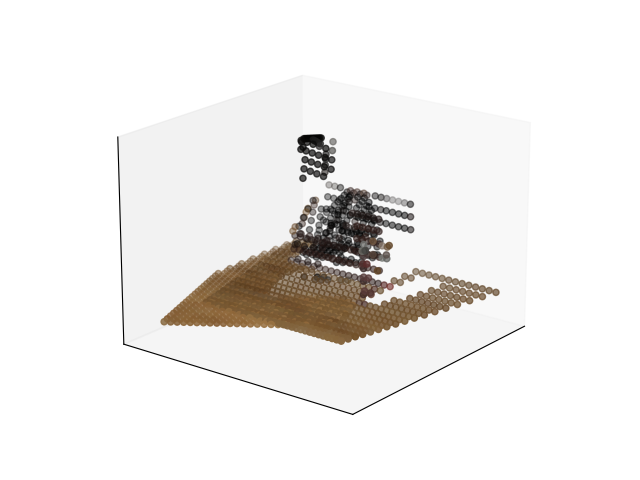}
        \caption{EE cropping}
        \label{fig:crop-hand}
    \end{subfigure}
    
    \caption{In this figure we present visualizations of the various cropping schemes discussed earlier in the paper. (a) The original RGB for reference. (b) The uncropped point cloud. The background is visible behind the robot. (c) Loose cropping. The robot is visible but the background is not. (d) Tight cropping. The table is fully visible but the robot base is not. (e) Tight cropping + hand cropping. We use the robot's proprioception to remove all points behind the end effector.}
    \label{fig:crop}
\end{figure*}

\begin{table}[!h]
\centering
\caption{The effect of world frame cropping on \algabbr{}. `None' means no cropping, `Loose' means that the background is cropped out but objects near the table, such as the robot, are not cropped out (see Figure~\ref{fig:crop-loose}). `Tight' means that nearer objects such as the robot base are cropped out. `No EE' omits the end effector cropping discussed in Section~\ref{sec:attention}, and all other experiments use it. All experiments are from combining \algabbr{} with BAKU.}
\vspace{-0.05in}
\label{table:cropping}

\begin{tabular}
{lcccc}
\toprule
 & None & Loose & Tight, No EE & Tight \\
\midrule
Orig. & $0.917$ & $0.914$ & $0.913$ & $\mathbf{0.919}$ \\
\midrule
UR5e & $0.772$ & $0.790$ & $0.778$ & $\mathbf{0.801}$ \\
Kinova3 & $0.648$ & $0.641$ & $\mathbf{0.674}$ & $0.653$ \\
IIWA & $0.695$ & $0.715$ & $0.700$ & $\mathbf{0.718}$ \\
\midrule
Small & $\mathbf{0.858}$ & $0.849$ & $0.834$ & $0.847$ \\
Medium & $0.830$ & $\mathbf{0.830}$ & $0.822$ & $0.829$ \\
Large & $0.802$ & $0.813$ & $0.821$ & $\mathbf{0.829}$ \\
\bottomrule
\end{tabular}

\end{table}

An important detail in several recent works is cropping, or the lack thereof \cite{3d_diffuser_actor, ze2024dp3, ze2024idp3}. In this experiment, we set out to understand how different cropping schemes impact the success rate of the final policy. Specifically, we compare the cropping schemes in Figure~\ref{fig:crop}, including:
\begin{itemize}
    \item No cropping
    \item Loose cropping, which removes the faraway background but not nearby objects such as the robot base.
    \item Tight cropping, which removes nearer objects such as the robot base but does not remove the table.
    \item EE cropping, which crops out points behind the end effector. 
\end{itemize}

Results are presented in Table~\ref{table:cropping}. We see that cropping leads to a modest improvement in in-distribution performance. Surprisingly it does not lead to a substantial improvement for the change of robot experiments, with the exception that end effector cropping leads to a modest improvement. We see that cropping does lead to a noticeable improvement for the camera change experiments. This makes sense since these experiments introduce many out of distribution faraway points that are subsequently cropped out. 

\subsubsection{Downsampling}

In this section we study the downsampling strategies discussed in Section~\ref{sec:attention}. Specifically, we compare the position-based downsampling used by DP3 \cite{ze2024dp3} and iDP3 \cite{ze2024idp3} to downsampling based on image features in $F$, as in \cite{3d_diffuser_actor}. 

First, we compare quantitative results in Table~\ref{table:downsampling}. We see that feature-based downsampling leads to strong improvements across the board, especially when generalizing to novel camera viewpoints. This is likely because changing camera positions introduces a large number of out of distribution points on the table, and downsampling according to features, which removes many repetitive points with the same features corresponding to points on the table, helps to mitigate their influence on the final policy.
\begin{table}[!h]
\centering
\caption{How does the choice of metric space for downsampling affect downstream performance? In this table we compare performance with different choices of metric space for farthest point sampling. We see that sampling according to 
$\ell_2$ distance between features in $F$ (right) outperforms sampling based on distances between Cartesian coordinates (left). Based on combining \algabbr{} with BAKU.}
\label{table:downsampling}

\begin{tabular}
{lcccc}
\toprule
 & Position-Based FPS & Feature-Based FPS \\
\midrule
Orig. & $0.916$ & $\mathbf{0.919}$ \\
\midrule
UR5e & $0.768$ & $\mathbf{0.801}$ \\
Kinova3 & $0.596$ & $\mathbf{0.653}$ \\
IIWA & $0.642$ & $\mathbf{0.718}$ \\
\midrule
Small & $0.717$ & $\mathbf{0.847}$ \\
Medium & $0.651$ & $\mathbf{0.829}$ \\
Large & $0.709$ & $\mathbf{0.829}$ \\
\bottomrule
\end{tabular}

\vspace{-0.1in}
\end{table}

\begin{figure*}
    \centering
    \begin{subfigure}[b]{0.4\textwidth}
        \centering
        \includegraphics[width=0.7\textwidth]{figures/images/cropping/rgb.png}
        \caption{RGB Image}
        \label{fig:downsample-rgb}
    \end{subfigure}
    \begin{subfigure}[b]{0.4\textwidth}
        \centering
        \includegraphics[width=\textwidth]{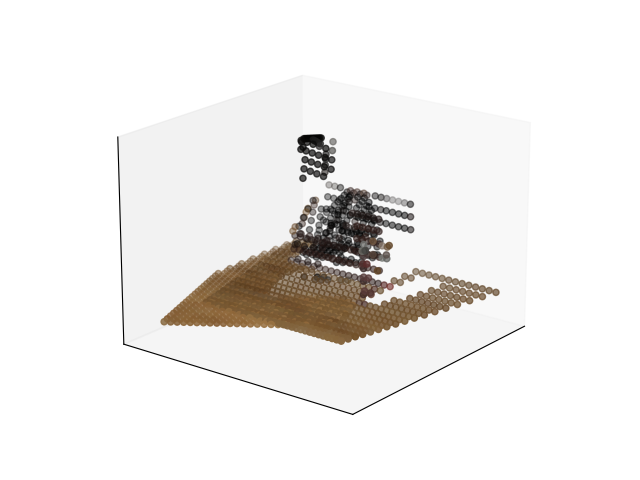}
        \caption{No downsampling}
        \label{fig:downsample-orig}
    \end{subfigure}
    \begin{subfigure}[b]{0.4\textwidth}
        \centering
        \includegraphics[width=\textwidth]{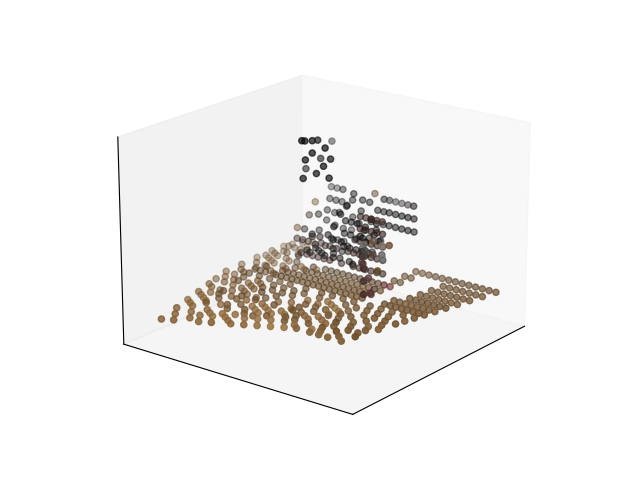}
        \caption{Downsampling based on location}
        \label{fig:downsample-pos}
    \end{subfigure}
    \begin{subfigure}[b]{0.4\textwidth}
        \centering
        \includegraphics[width=\textwidth]{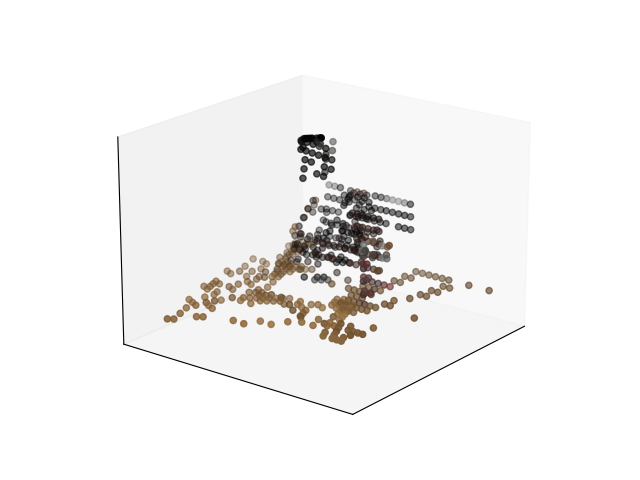}
        \caption{Downsampling based on image features $F$}
        \label{fig:downsample-feat}
    \end{subfigure}

    \caption{In this figure we present visualizations of the various downsampling schemes discussed earlier in the paper. (a) RGB image of the scene for reference. (b) The original point cloud after cropping. (c) The point cloud after downsampling based on Cartesian coordinates. (d) Downsampling based on image features $F$.}
    \label{fig:downsampling}
\end{figure*}

Next, we study a qualitative comparison of the two strategies in Figure~\ref{fig:downsampling}. We see in Figure~\ref{fig:downsample-pos} that position-based downsampling samples many points on the table, whereas in Figure~\ref{fig:downsample-feat}, the points are concentrated more around the objects in the scene.

Finally, we study the effect of $p$, the hyperparameter controlling the number of points in the downsampled point cloud, in Table~\ref{table:downsampling_p}. We see that it is not particularly sensitive to this hyperparameter, with a slight improvement for $p=512$.

\begin{table}[]
\centering
\caption{In this table we vary the number of points in the downsampled point cloud, $p$, when combining \algabbr{} with BAKU. We see that while performance is not particularly sensitive to this hyperparameter, there is slightly better performance for $p \in \{256, 512\}$.}
\label{table:downsampling_p}

\begin{tabular}
{lcccc}
\toprule
 & $p=128$ & $p=256$ & $p=512$ & $p=1024$ \\
\midrule
Orig. & $0.917$ & $\mathbf{0.919}$ & $\mathbf{0.919}$ & $0.918$ \\
\midrule
UR5e & $0.785$ & $\mathbf{0.802}$ & $0.801$ & $0.776$ \\
Kinova3 & $0.643$ & $\mathbf{0.681}$ & $0.653$ & $0.643$ \\
IIWA & $0.725$ & $\mathbf{0.745}$ & $0.718$ & $0.672$ \\
\midrule
Small & $0.828$ & $0.843$ & $\mathbf{0.847}$ & $0.835$ \\
Medium & $0.802$ & $0.804$ & $\mathbf{0.829}$ & $0.803$ \\
Large & $0.814$ & $0.804$ & $\mathbf{0.829}$ & $0.799$ \\
\bottomrule
\end{tabular}

\end{table}

\subsubsection{Proprioception}

In Table~\ref{table:proprioception} we study the effect of proprioceptive inputs on the final LIBERO results. The logic in removing it is that, because the point cloud is in the end effector's coordinate frame, the proprioceptive information is arguably baked into the observation. We see that removing the proprioception leads to a modest benefit for in-distribution and change of embodiment experiments, but leads to a massive degradation in performance when changing camera poses. Thus, for our final model we include proprioception.

\begin{table*}[]
\centering
\caption{with and without proprioception}
\vspace{-0.05in}
\label{table:proprioception}
\begin{tabular}
{l|c|ccc|ccc}
\toprule
Algorithm & Orig. & UR5e & Kinova3 & IIWA & Small & Medium & Large \\
\midrule
ACT + \algabbr{} without & $\mathbf{0.916}$ & $\mathbf{0.824}$ & $\mathbf{0.796}$ & $\mathbf{0.760}$ & $0.671$ & $0.472$ & $0.452$ \\
ACT + \algabbr{} with & $0.912$ & $0.789$ & $0.773$ & $0.731$ & $\mathbf{0.779}$ & $\mathbf{0.702}$ & $\mathbf{0.760}$ \\
\midrule
DP + \algabbr{} without & $\mathbf{0.899}$ & $\mathbf{0.761}$ & $\mathbf{0.568}$ & $\mathbf{0.522}$ & $0.568$ & $0.419$ & $0.398$ \\
DP + \algabbr{} with & $0.896$ & $0.737$ & $0.528$ & $0.488$ & $\mathbf{0.783}$ & $\mathbf{0.794}$ & $\mathbf{0.782}$ \\
\midrule
BAKU + \algabbr{} without & $\mathbf{0.931}$ & $\mathbf{0.813}$ & $\mathbf{0.757}$ & $0.696$ & $0.747$ & $0.601$ & $0.570$ \\
BAKU + \algabbr{} with & $0.920$ & $0.787$ & $0.654$ & $\mathbf{0.701}$ & $\mathbf{0.854}$ & $\mathbf{0.831}$ & $\mathbf{0.840}$ \\
\bottomrule
\end{tabular}

\end{table*}

\end{appendix}

\end{document}